\pdfoutput=1

\documentclass[11pt, dvipsnames]{article}

\usepackage{acl}

\usepackage{times}
\usepackage{latexsym}

\usepackage[T1]{fontenc}

\usepackage[utf8]{inputenc}

\usepackage{microtype}

\usepackage{inconsolata}

\usepackage{placeins}
\usepackage{comment}
\usepackage{graphicx}
\usepackage{tabularx}
\usepackage{soul}
\usepackage{multirow}
\usepackage{multicol}
\usepackage{colortbl}
\usepackage{tikz}
\usepackage{amsmath}
\usepackage{hyperref}
\usepackage{euflag}
\makeatletter
\newcommand{\@final}{2}

\newcommand{\@todobase}[3]{\colorbox{#1}{\textbf{#2}}\hspace{5pt}\textcolor{#1}{#3}}
\newcommand{\@todo}[2][red]{\@todobase{#1}{ToDo:}{#2}}
\newcommand{\@todostar}[3][red]{\@todobase{#1}{#2}{#3}}
\newcommand{\todo}{\@ifstar{\@todostar}{\@todo}}

\newcommand{\@rvplus}[3]{
\if\@final0%
\@todobase{#1}{#2}{#3}%
\else%
#3
\fi%
}
\newcommand{\plus}[2]{\@rvplus{green}{#1}{#2}}

\newcommand{\@rvminus}[3]{
\if\@final0%
\@todobase{#1}{#2}{\sout{#3}}%
\fi%
}
\newcommand{\minus}[2]{\@rvminus{red}{#1}{#2}}

\newcommand{\anonymous}[2][\empty]{
\if\@final1%
#1%
\else%
#2%
\fi%
}

\makeatother

\graphicspath{{./pictures/}}

\title{CICLe: Conformal In-Context Learning for Largescale Multi-Class Food Risk Classification}

\author{%
Korbinian Randl,$^{1}$
John Pavlopoulos,$^{1,2,3}$
Aron Henriksson,$^{1}$
Tony Lindgren$^{1}$
\\
$^{1}$ Stockholm University, 
Borgarfjordsgatan 12, 164 07 Kista, Sweden\\
\texttt{\{korbinian.randl,ioannis,aronhen,tony\}@dsv.su.se}\\
$^{2}$Athens University of Economics and Business, Patission 76, Athens 104 34, Greece\\
$^{3}$Archimedes/Athena RC
}

\begin{document}
\maketitle
\begin{abstract}
Contaminated or adulterated food poses a substantial risk to human health. Given sets of labeled web texts for training, Machine Learning and Natural Language Processing can be applied to automatically detect such risks.
We publish a dataset of 7,546 short texts describing public food recall announcements. Each text is manually labeled, on two granularity levels (coarse and fine), for food products and hazards that the recall corresponds to. 
We describe the dataset and benchmark naive, traditional, and Transformer models. Based on our analysis, Logistic Regression with a tf-idf representation outperforms RoBERTa and XLM-R on classes with low support.
Finally, we discuss different prompting strategies and present an LLM-in-the-loop framework, based on Conformal Prediction, which boosts the performance of the base classifier while reducing energy consumption compared to normal prompting.
\end{abstract}

\section{Introduction}

Food-bourne illnesses and contaminated food pose a serious threat to human health and lead to thousands of deaths \cite{Majowicz2014_ecoli, Maertens_lListeriosis}.
Natural Language Processing~(NLP) solutions based on Machine and Deep Learning~(ML, DL) or Large Language Models (LLMs) enable fast responses to new threats by generating warnings from publicly available texts on the internet \cite{Maharana2019_DetectingReportsOfUnsafeFoods, Tao2023_FoodborneIllnessDetection}. These texts, however, as we show, can be noisy and are characterized by thousands of classes.
Categorization of incidents into these classes is needed to properly address food risks, as different food risks require different countermeasures with different urgency. For example, while a fraudulent import of foodstuffs usually imposes a very low risk to consumer health, Listeria or Salmonella contamination is a serious risk to consumer health and can even lead to death. Automatic classification of the hazard involved in food recalls can help estimate the severity of food risks and lead to faster and more targeted measures.
While we are aware of existing text datasets on the topic of food-bourne illnesses \cite{hu2022_tweetfid}, they focus only on the detection of such illnesses and not on their categorization. Publicly available official food recall announcements, and specifically their shorter titles, contain detailed information on food risks and the specific products involved and may be used as a basis for data-driven ML approaches. 
Acknowledging this research gap, this paper is specifically addressing the following research questions:

\vspace{8pt}
\noindent\textbf{R1:} Are titles of food recall articles sufficient for classification of the food hazards and food products involved?
\begin{itemize}

    \vspace{-8pt}
    \item\footnotesize\textbf{R1.1:} Do the titles of official announcements of food hazards comprise enough information for their classification?

    \vspace{-8pt}
    \item\footnotesize\textbf{R1.2:} What is the performance of ML and DL classification models trained on the presented dataset?
\end{itemize}

\noindent As LLMs like GPT-4, and Llama~\cite{Touvron_llama} represent the current state-of-the-art in NLP, but are also known for their high resource consumption, we further explore the following question:

\vspace{8pt}
\noindent\textbf{R2:} Can LLMs be used to address or improve automatic food hazard classification?
\begin{itemize}

    \vspace{-8pt}
    \item\footnotesize\textbf{R2.1:} What is the performance of LLMs for the classification tasks? 

    \vspace{-8pt}
    \item\footnotesize\textbf{R2.2:} How can ML and DL be used to increase resource efficiency without harming the accuracy of LLMs for the task?
\end{itemize}

\noindent In this work, we address the above task of classifying food hazards and the involved products from food recall texts and provide the following main contributions:
\begin{enumerate}
    \item We present the first dataset for text classification of food products and food hazards on two levels of granularity.\footnote{
    \anonymous[\color{blue}{(hidden for reasons of anonymity)}]{\href{https://doi.org/10.5281/zenodo.10820657}{DOI:10.5281/zenodo.10820657}}}

    \item We present a benchmark on the introduced dataset using naive, traditional ML, and Transformer classifiers, showing that a Support Vector Machine outperforms the rest due to better performance on low-support classes.

    \vspace{-8pt}
    \item We discuss different approaches to few-shot prompting with GPT-3.5 on our data and propose reducing the number of relevant classes for few-shot prompting using Conformal Prediction~(CP). We show that this enables resource-friendly prompting at comparably high $F_1$ by boosting a base classifier's performance in a LLM-in-the-loop framework. 

\end{enumerate}

\section{Background and Related Work}\label{sec:related_work}

\paragraph{Research on ML with food data} has traditionally focused on tabular and image data \cite{Zhou2019_DeepLearningInFood, Jin2020_BigDataInFoodSafety, Wang2022_MLforFoodSafety}, where food risk classification is usually presented as a detection task: rather than identifying specific hazards, the task is formulated as the binary classification of the presence or absence of an incident \cite{Maharana2019_DetectingReportsOfUnsafeFoods, Tao2023_FoodborneIllnessDetection, Wang2023_BayesianNetworkForFoodSafety}.
Focusing on the text modality, we found literature exploring such detection tasks based on Amazon reviews matched with FDA food recall announcements~\cite{Maharana2019_DetectingReportsOfUnsafeFoods}, as well as Twitter data, labeled using instances from the US National Outbreak Reporting System~\cite{Tao2023_FoodborneIllnessDetection}. In both of these papers, the data is highly unbalanced towards the negative class (i.e. no incident). Furthermore, the description of the method for labeling the Twitter texts in \citet{Tao2023_FoodborneIllnessDetection} lacks important details, as it is unclear on what grounds texts and events are matched.
We only found one study on the actual classification of hazards: \citet{Xiong2023_FoodSafetyNewsEventsClassification} which uses a hierarchical Transformer model to classify Chinese news texts into four categories of hazards. The authors do not share further details on their data, which makes an assessment of the paper's validity hard.
Overall, the field of food risk classification based on text lacks reproducible work on fine-grained prediction of hazards based on openly accessible data.

\paragraph{ML models based on the attention mechanism} by \citet{Vaswani2017_AttentionIsAllYouNeed} and the self-supervised masked language modeling pre-training paradigm generally outperform traditional methods both on monolingual~\cite{Devlin_bert, Liu_roberta} and multilingual~\cite{Conneau_xlm-r} data.
Although such Transformers were originally described as an encoder-decoder architecture \cite{Vaswani2017_AttentionIsAllYouNeed}, mapping an input-text to an output-text, models intended for classification tasks usually only employ an encoder \cite{Devlin_bert, Liu_roberta, Conneau_xlm-r}, followed by several task-specific layers.

Recently, LLMs such as Llama~2~\cite{Touvron_llama}, PaLM~\cite{Chowdhery_palm}, and GPT-family models~\cite{Brown2020_FewShotLearning} have been shown to exceed the capabilities of smaller Transformers even without further fine-tuning. That is, by simply providing a context of a few labeled samples per class, LLMs can predict the classes of unseen samples within this context. As LLMs are usually text-to-text Transformers, the context is provided directly in each prompt. This paradigm is commonly referred to as \textit{few-shot prompting} or \textit{in-context learning} \cite{Brown2020_FewShotLearning}. Since the discovery of the few-shot capabilities of LLMs, newer LLMs are often designed for high few-shot performance \cite{Gao2021_BetterFewShotLearners, Chowdhery_palm}.

\paragraph{Current work on few-shot prompting} focuses mostly on creating/finding the optimal few-shot samples from the training data.
\citet{Ahmed2023_FewShot+StaticAnalysis} proposed a workflow for automatically finding similar samples from the pool of labeled ``training'' samples for code summarization. \citet{Shi2023_PromptSpace} focused on automatic generation of Chain of Thought~(CoT) labels for samples in reasoning tasks. CoT is another prompting paradigm that asks the LLM to provide a chain of reasoning before delivering the prediction and has been shown to drastically improve reasoning performance \cite{Wei2022_CoT}. Nevertheless, to the best of our knowledge, there is no previous work on how to leverage few-shot prompting in LLMs for multi-class prediction problems with a high number of classes. While prompting with long contexts may ``confuse'' the LLM \citep{liu_contexts}, the context window may even be too small to support a view of all classes in such cases. Furthermore, the previous approaches focus on selecting specific samples rather than relevant classes for the prompt.

\paragraph{The CP framework}\cite{Vovk_conformal} is a method for associating predictions of a classification algorithm with confidence guarantees. Essentially, CP selects sets of classes from the predicted labels of a classifier or regressor that statistically contain the true label with a predefined probability. 
We note, however, that the standard CP setting provides guarantees for the individual class labels; i.e., errors for one of the labels can, hence, exceed the set guarantees. A formulation, called Mondrian of CP, can make the guarantees hold also for the individual labels. Note that CP does a very good job of keeping the output (set of labels) small, while the prediction guarantees hold. 
It can be applied to any classification (or regression) model, as long as a calibration set is set aside together with a measure of the model's prediction error with regard to the true labels. This measure is referred to as the ``\textit{non-conformity function}''. In our context, the stronger the model, and the better the non-conformity function, the fewer labels (hence, shots) will be produced with the same guarantee. We point to the work of \citet{Vovk_conformal, johansson2014regression, 8260735} for more information on CP.

\section{The Food Recall Dataset}\label{sec:data}
In this section, we first describe the dataset used and shared in this paper, then discuss the labeling process and quality, as well as class distribution.

\subsection{Data Description}
\begin{figure}[!ht]
\begin{center}
\includegraphics[width=\linewidth]{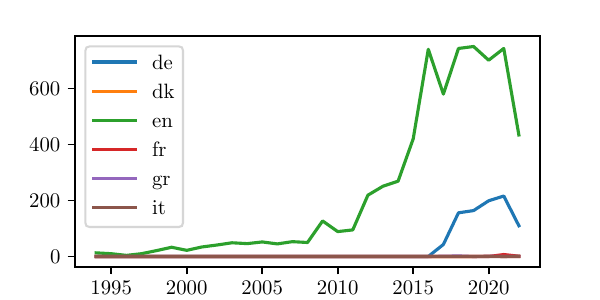} 
\caption{Languages in the dataset per year.}
\label{fig:language_per_year}\vspace{-15pt}
\end{center}
\end{figure}

The dataset consists of 7,546 short texts (length in characters: min=5, avg=84, max=360), which are the titles of food-recall announcements (therefore referred to as \texttt{title}), crawled from 24 food-recall domains (governmental \& NGO, see Table~\ref{tab:domains}, Appendix~\ref{sec:appendix_data}) by Agroknow\footnote{\url{https://agroknow.com/}} \cite{Agroknow_crawler}. 
The texts are written in 6 languages, with English ($n = 6,644$) and German ($n = 888$) being the most common, followed by French ($n = 8$), Greek ($n = 4$), Italian ($n = 1$) and Danish ($n = 1$). As shown in Figure~\ref{fig:language_per_year}, most of the texts have been authored after 2010.
The texts describe recalls of specific food products due to specific reasons. Experts manually classified each text to four groups of classes describing hazards and products on two levels of granularity:

\noindent\textbf{\texttt{hazard}}: A fine-grained description of the hazards mentioned in the texts comprising $261$~classes.
\noindent\textbf{\texttt{hazard-category}}: A categorized version of the \texttt{hazard} classification task comprising $10$~classes.
\noindent\textbf{\texttt{product}}: A fine-grained description of the products mentioned in the texts comprising 1,256 classes.
\noindent\textbf{\texttt{product-category}}: A categorized version of the \texttt{product} classification task comprising $22$~classes.
The columns \texttt{hazard-title} and \texttt{product-title} comprise character spans, generated based on feature importance of a Logistic Regression (LR) classifier (see §\ref{sec:ground_truth}). These signify parts of the \texttt{title} that are important for the hazard and product classification.
As the fine-grained tasks, \texttt{hazard} and \texttt{product}, have very low support for many classes, they may require further pre-processing (e.g. clustering/filtering of labels) depending on the application.
The dataset, publicly released under a \href{https://creativecommons.org/licenses/by-nc-sa/4.0/deed.en}{Creative Commons BY-NC-SA~4.0} license, comprises also metadata, such as the release date of the text (columns \texttt{year}, \texttt{month}, and \texttt{day}), the language of the text (column \texttt{language}), and the country of issue (column \texttt{country}). Appendix~\ref{sec:appendix_data} contains more detailed statistics, as well as sample texts.

\subsubsection{Ground Truth}\label{sec:ground_truth}
It is important to note that samples were labeled not only based on the \texttt{title}, but also the content of the food recall article. This means that some of the samples may not contain evidence for all the classes assigned to them. As samples that are missing specific information are common in real-world data, we decided not to filter out such samples; instead, we provide an estimate of such noise in the data in the \texttt{hazard-title} and \texttt{product-title} classification tasks.
To that end, we use the coefficients of the tf-idf Logistic Regression (TF-IDF-LR) classifiers for the \texttt{hazard-category} and \texttt{product-category} classes to extract important terms per class. 

For each text-label pair~$T_i, Y_i$, we split $T_i$ in tokens $\{t_{i,1}, t_{i,2}, \dots, t_{i,K}\}$, using the process described in \S\ref{sec:classic_ml}. We then calculate a score by adding the positive model coefficients associated with $t_{i,k}$ if $y_{i,j} = 1$, and subtracting the positive model coefficients associated with $t_{i,k}$ if $y_{i,j} = 0$. 
Although the quality of these terms depends on class support, they can still help us frame the noise in the data by focusing on informative tokens, i.e., tokens with a positive coefficient for a specific class. We find that each such token corresponds to $1.16$ classes on average for the \texttt{hazard-category} and $1.30$ classes for the \texttt{product-category}.\footnote{Estimation for the fine-grained classification tasks is difficult because of low per-class support.}
Also, we see that $13.1\%$ of the samples do not have spans in \texttt{hazard-title} ($17.1\%$ for \texttt{product-title}), indicating that evidence for the class is missing.

Fine-grained labels were assigned per web-domain by a domain expert of Agroknow and then grouped to the final labels using internal ontologies.\footnote{Random checks of the labels were performed by more experienced curators for quality assurance.}
In order to measure inter-annotator agreement~(IAA), we asked a senior curator involved in the labeling process and one independent expert to label a sample of 30 texts (stratified over both \texttt{*-category} labels) independently of each other. Comparing the annotator's assessment to the labels in the dataset shows moderate agreement (Cohen's~$\kappa$: 0.76 on \texttt{hazard-category}, 0.63 on \texttt{product-category}). The expert's assessment is intuitively slightly less aligned (Cohen's~$\kappa$: 0.63 on \texttt{hazard-category}, 0.60 on \texttt{product-category}) but still acceptable. Expert and Annotator show a $\kappa$ of 0.59 on \texttt{hazard-category}, 0.77 on \texttt{product-category}. The raw data of our IAA study are available along with the code.

\begin{figure}[!t]
\begin{center}
\includegraphics[width=\linewidth]{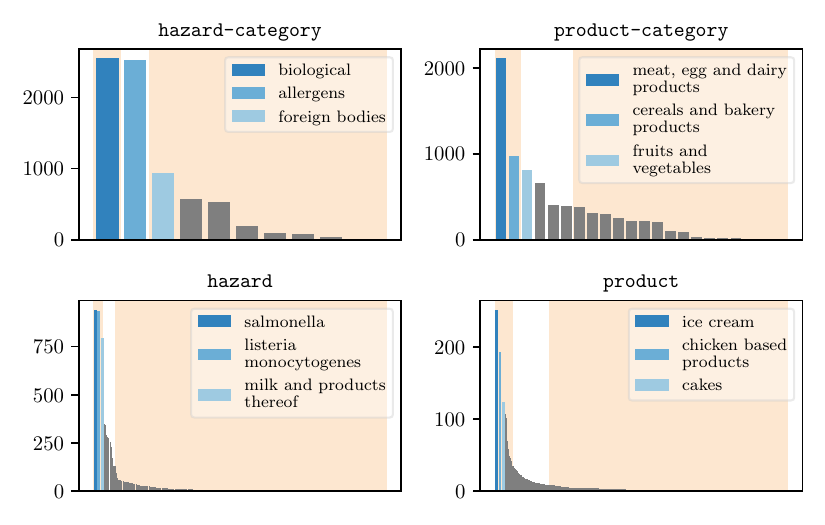}
\caption{Class support with orange background for high- (on the left) and low-support classes (right) for performance analysis. For reasons of better readability, we only name the three most supported classes per task.}
\label{fig:label_distribution}\vspace{-20pt}
\end{center}
\end{figure}

\subsection{Class Imbalance}
One of the most prominent features of the data is the heavy class imbalance. Figure~\ref{fig:label_distribution} shows the sample counts per class and classification task. All the classification tasks in the data show a long-tail distribution of classes, with just a small number of classes having most of the samples. Therefore, we extract sets of high-support classes $\mathcal{C}_{high}$ and low-support classes $\mathcal{C}_{low}$ comprised of around one-third of the total number of samples in the data for each classification task. The classes included in these sets are highlighted by an orange background in Figure~\ref{fig:label_distribution}. For the \texttt{hazard-category} classification task, the $\mathcal{C}_{high}$ is comprised of only one class with $2,557$~samples, and the $\mathcal{C}_{low}$ consists of $8$~classes with $2,462$~samples in total, and for \texttt{hazard} $|\mathcal{C}_{high}|=2,668$~samples in $3$~classes and $|\mathcal{C}_{low}|=2,414$~samples in $248$~classes.
For \texttt{product-category} we have $|\mathcal{C}_{high}|=3,089$~samples in $2$~classes and $|\mathcal{C}_{low}|=2,185$~samples in $16$~classes, and for \texttt{product} $|\mathcal{C}_{high}|=2,508$~samples in $39$~classes and $|\mathcal{C}_{low}|=2,483$~samples in $1,058$~classes.

\section{Methodology}\label{sec:method}
This section describes the problem setting and the methods utilized in our empirical evaluation.

\subsection{The Task}
The basic problem we address in this paper is extreme multi-class classification on heavily imbalanced data. More formally, given a number of training texts~$T_i,~i\in\{1,2,\dots, N\}$ and their corresponding classes~$y_i$, we aim to train a classifier~$\mathrm{f}(T_i,\cdot)=\hat{y}_i$ that minimizes the error $|y_i-\hat{y}_i|$. For standard ML classifiers, the function~$\mathrm{f}(T_i, \cdot)$ is usually a two-step process with the first step mapping the text to a machine-readable embedding vector~$X_i = \mathrm{e}(T_i)$, and the second one involving the learning process on $X_i$: $\mathrm{f}(T_i,\cdot)=\mathrm{f'}(X_i,\cdot)$.
Both the class labels~$y_i$ and predictions~$\hat{y}_i$ are integer scalars in $\{1,2,\dots, M\}$. In all our classification tasks, we have at least $M\ge10$.

\subsection{Classifiers}
As a naive baseline, we report the performance of two classifiers. The random classifier~(RANDOM) yields a random integer in $\{1,2,\dots, M\}$ for each $\hat{y}_{i}$. 
The majority baseline~(MAJORITY) chooses each $\hat{y}_{i}$ to be the class with the highest number of samples in the training data.

\subsubsection{Traditional ML Classifiers}\label{sec:classic_ml}
We use Bag-of-Words~(BOW) and Term Frequency - Inverse Document Frequency~\cite[TF-IDF]{SpärckJones_idf} encodings (see Appendix~\ref{sec:appendix_models}), combined with $k$-Nearest Neighbours~(KNN), LR, or Support Vector Machine~(SVM) classifiers.\footnote{For the classifiers, we use the implementation from the Python library \texttt{scikit-learn}~\cite{scikit-learn}.} For KNN, we optimize over $k\in\{2, 4, 8\}$ neighbours based on cosine similarity of the embedding vectors. For LR, we use a `liblinear' solver and optimize $C \in \{0.5, 1.0, 2.0\}$ for both L1 and L2 regularization. For the SVMs, we use a linear kernel and optimize the parameter $C \in \{0.5, 1.0, 2.0\}$ for L2 regularization on the validation splits (defined in §\ref{sec:setup}). For LR and SVM, we use a one-vs-all approach to classification, which means that we train one binary classifier for each class in the predicted classification task and choose the class with the highest probability.

\subsubsection{Encoder-only Transformers}
As a more recent counterpart to the previously described traditional ML classifiers, we fine-tune two models from huggingface's \texttt{Transformers}\footnote{\href{https://huggingface.co/docs/transformers/index}{https://huggingface.co/docs/transformers/index}} library: RoBERTa$_{base}$~\cite{Liu_roberta} and XLM-RoBERTa$_{base}$~\cite[XLM-R]{Conneau_xlm-r} in their \texttt{base}-sizes (RoBERTa: \textit{125M params}; XLM-R: \textit{270M params}). Both models use the structure introduced by BERT$_{base}$~\cite[\textit{L=12, H=768, A=12}]{Devlin_bert}, which improves comparability of their results. The different parameter counts result mainly from the different vocabulary sizes used in their Byte-Pair-based encoders~\cite{Sennrich_bpe}: RoBERTa uses a vocabulary of size 50k, while XLM-R uses 250k tokens. For the purposes of this paper, the most important difference between the models is that while RoBERTa is only pre-trained on English texts, XLM-R is pre-trained on 100 different languages.

To fine-tune these two models, we use the standard sequence classification heads provided by the \texttt{Transformers} library. We optimize training using AdamW~\cite{Loshchilov_adam-w} in combination with a learning rate that stays constant at a value of $5\cdot10^{-5}$ for the first two epochs and then declines linearly towards 1\% of its starting value after 20 epochs. Furthermore, we employ early stopping with a patience of five epochs on the minimum cross-entropy loss computed on the validation set. Due to hardware limitations, we use a batch size of 16 observations.

\subsection{Prompting}
Prompt engineering can lead to accurate classification results with limited training data.
We investigate this path for our task by employing few-shot in-context learning with \texttt{gpt-3.5-turbo-instruct}.\footnote{Accessed on 02-2024 through OpenAI's Python API.} For all our prompting examples, we use a temperature value of $0$.
The naive approach is to provide a context describing the classification task followed by the two most similar text-label pairs (based on cosine similarity of the TF-IDF-embeddings) per class from the training data. These examples are ordered from the most similar to least similar in the prompt (dubbed \textbf{GPT-ALL}).

\begin{figure}[th]
\begin{center}
\includegraphics[width=\linewidth]{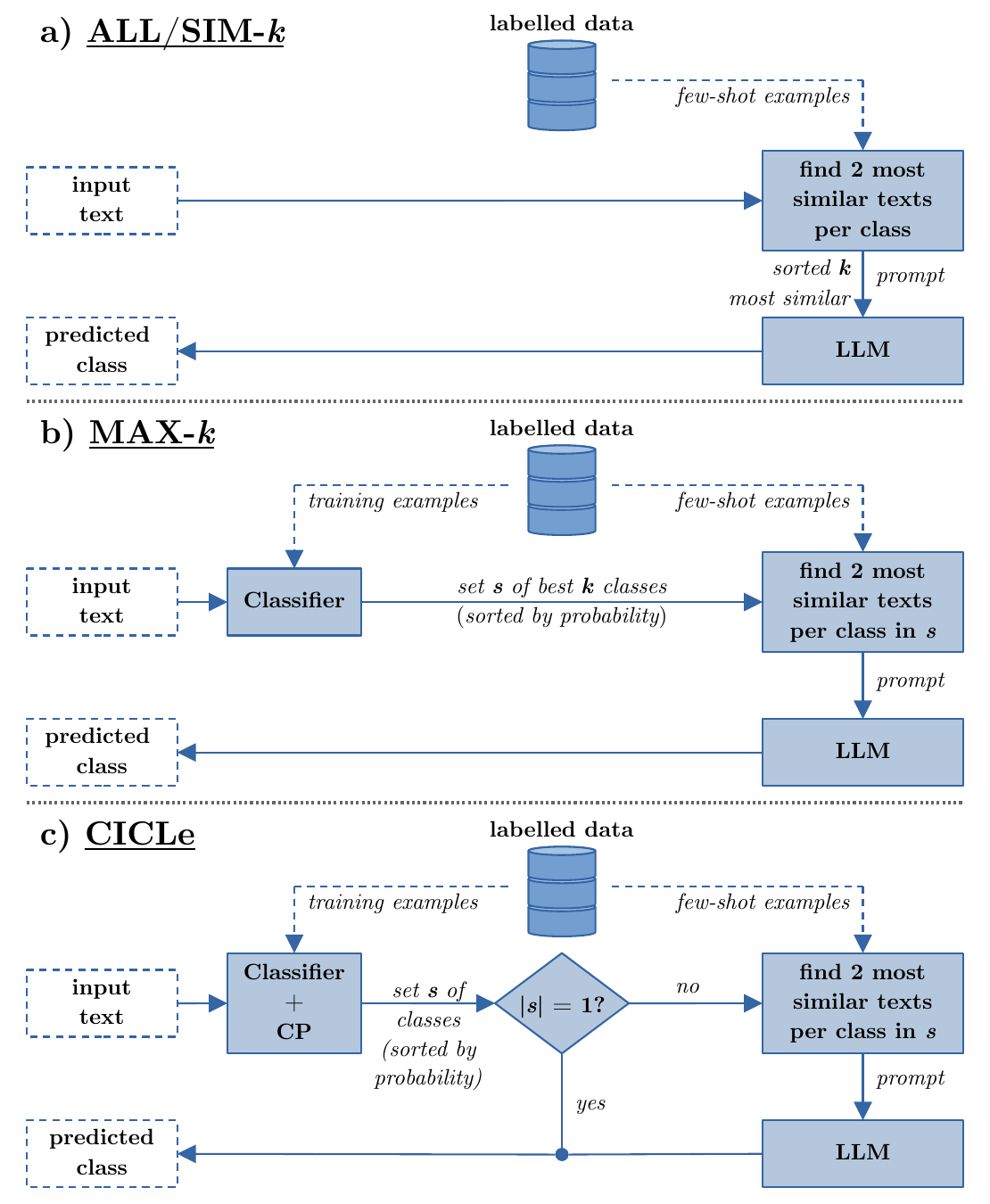}
\caption{Prompting strategies used in this study. Although the strategies are generally independent of the choice of classifier and LLM, we only show results for TF-IDF-LR and OpenAI's \texttt{gpt-3.5-turbo-instruct}.}
\label{fig:prompting_uml}\vspace{-10pt}
\end{center}
\end{figure}

GPT-ALL is applicable only to coarse tasks due to the size of the context window: for the \texttt{hazard} task, we would have at most $\frac{4,096~\mathrm{tokens}}{261~\mathrm{classes}} = 15.7$ tokens per class which is insufficient for providing two example texts. Furthermore, reduced prompt sizes can lead to increased energy efficiency, as the size of the attention weights computed in the Transformer is \textit{input~size}~$\times$~\textit{output~size} \citep{Vaswani2017_AttentionIsAllYouNeed}. Lastly, meaningfully reducing the classes in the prompt can increase predictive performance, as the probability of choosing the right class increases (Appendix~\ref{sec:appendix_results}).
Therefore, we use three different kinds of prompts (see Figure~\ref{fig:prompting_uml}) with the third kind (GPT-CICLe) being novel. \textbf{GPT-SIM-$k$} is simply the GPT-ALL prompt shortened by only keeping the $k$ most similar few-shot samples.
\textbf{GPT-MAX-$k$} selects the two most similar few-shot samples from the $k$ most probable class labels as predicted by a base classifier. The few-shot samples are ordered from most to least probable, which yields $2 \cdot k$ few-shot samples per prompt.

\subsubsection{Our proposed GPT-CICLe}
We consider that the better the few shots the LLM is provided with, the better its performance will be. Therefore, 
in this work, we propose using Conformal Prediction (CP)~\cite{Vovk_conformal} to adapt the context length to a base classifier's certainty on the predicted classes. 
CP uses the concept of a non-conformity measure $\delta(y_i, \hat{y}_i)$ to create sets of predicted classes, which contain the true class with a probability of $p \approx 1-\alpha$. 

In our case, this non-conformity measure is simply the base classifier's uncertainty on the true class: \[\delta(y_i, \hat{y}_i) = 1 - P(\hat{y}_{i} = y_{i}).\]
Here, $\hat{y}_i$ are the predictions and $y_i$ are the corresponding true labels.
If one defines \[q=\frac{\lceil{(N+1)(1-\alpha)}\rceil}{N},\] and $\hat{q}$ as the $q^{th}$ empirical quantile of $\{\delta(Y_i, \hat{Y}_i) | 1 \le i \le N\}$, it can be shown that for any prediction $\hat{Y}$ of the classifier on an unknown sample, a set of classes $\{j | \hat{y}_{j} \ge 1-\hat{q}\}$ contains the true class with probability $p \ge 1-\alpha$~\cite{Vovk_conformal}.
This means that we accept an upper bound of $1-\alpha$ on prompting accuracy, which we trade for adaptive context length based on sample difficulty. 

In the extreme case of a single class in the prediction set, we bypass the LLM completely and predict this class.
As it is possible for CP to predict empty sets, which is not useful in classification, we return a set containing only the most probable class in such cases. This means that for $\alpha > 1-$\textit{base classifier accuracy}, the performance of this prompt becomes equal to that of the base classifier. 
As in GPT-MAX-$k$, we finally select the two most similar samples per class in the conformal set from the training data to compose the prompt. Samples are again ordered from most to least probable as determined by the classifier. We refer to this LLM-in-the-loop framework as \emph{Conformal In-Context Learning}~(CICLe).
An example of each of these prompts is shown in Appendix~\ref{sec:appendix_prompts}.

In this paper, we focus on the pure few-shot performance of \texttt{gpt-3.5-turbo-instruct}. Therefore, we decided not to employ additional prompt-engineering techniques, such as CoT. Our code is publicly available (under a \href{https://www.gnu.org/licenses/gpl-3.0.en.html}{GPL V3.0} license) at \anonymous[\href{https://anonymous.4open.science/r/conformal\_prompting-2E0B}{Anonymized GitHub repository}]{\href{https://github.com/k-randl/conformal\_prompting}{https://github.com/k-randl/conformal\_prompting}}.

\section{Empirical Analysis}\label{sec:analysis}

In this section, we present the experimental setup and the experimental results.

\vspace{-5pt}
\subsection{Experimental Setup}\label{sec:setup}
For training and evaluation of our ML models, we apply 5-fold cross-validation~(CV) to create 5 train-test splits. From each of these 5 training sets, we create a validation set using $10\%$ holdout which we use for hyperparameter tuning as described in Appendix~\ref{sec:appendix_models}.
To reduce the computational cost, we only test our prompting methods other than GPT-ALL on the first CV split, which corresponds to a $20\%$ holdout.
We use stratification on the \texttt{*-category} classification tasks as those provide a sufficient number of samples for splitting in each class. We use the same splits for \texttt{hazard-category} and \texttt{hazard}, as well as \texttt{product-category} and \texttt{product} to keep the results comparable.
Our classifiers provide baseline performance on the dataset. In order to demonstrate the effect of class imbalance on performance, we do not employ balancing methods like oversampling or class weights during training.

\vspace{-5pt}
\subsection{Experimental Results}
In this section, we present the predictive performance of the classifiers described in §\ref{sec:method}. For \texttt{hazard} and \texttt{product} we present performance on classes that have a minimum support of four in the train set and at least one instance in the test set as for these ML is feasible. Performance on all classes in these tasks is reported in Appendix~\ref{sec:appendix_results}.

\begin{table}[!ht]
\begin{center}
\resizebox*{\linewidth}{!}{
\begin{tabular}{|r|rlrlrl|}

    \hline
    \cellcolor{gray!25} &
    \multicolumn{2}{c}{\cellcolor{gray!25} $\bf F_1$ (\textit{\footnotesize all classes})} &
    \multicolumn{2}{c}{\cellcolor{gray!25} $\bf F_1$ ({\footnotesize $\mathcal{C}_{high}$})} &
    \multicolumn{2}{c|}{\cellcolor{gray!25} $\bf F_1$ ({\footnotesize $\mathcal{C}_{low}$})} \\

    \multicolumn{1}{|c|}{\cellcolor{gray!25} \multirow{-2}{*}{\textbf{Model}}} &
    \cellcolor{gray!25} {\footnotesize mean} &
    \cellcolor{gray!25} {\footnotesize max.} &
    \cellcolor{gray!25} {\footnotesize mean} &
    \cellcolor{gray!25} {\footnotesize max.} &
    \cellcolor{gray!25} {\footnotesize mean} &
    \cellcolor{gray!25} {\footnotesize max.} \\
    
    \hline
    \hline

    \multicolumn{2}{|l}{\cellcolor{gray!10}\footnotesize\texttt{hazard-category}} & 
    \multicolumn{5}{r|}{\cellcolor{gray!10}\footnotesize\textit{$10$ classes}} \\

    \hline

    RANDOM &
    \footnotesize $0.08$ & \footnotesize $0.08$ &
    \footnotesize{\color{gray} $0.33$} & \footnotesize{\color{gray}  $0.34$} &
    \footnotesize $0.00$ & \footnotesize $0.00$ \\
    
    MAJORITY &
    \footnotesize $0.05$ & \footnotesize $0.05$ &
    \cellcolor{blue!15}\footnotesize{\color{gray} $\bf 1.00$} & \cellcolor{blue!15}\footnotesize{\color{gray} $\bf 1.00$} &
    \footnotesize $0.00$ & \footnotesize $0.00$ \\

    \hline
    
    TF-IDF-KNN &
    \footnotesize $0.47$ & \footnotesize $0.50$ &
    \footnotesize{\color{gray} $0.45$} & \footnotesize{\color{gray}  $0.46$} &
    \footnotesize $0.41$ & \footnotesize $0.46$ \\
    
    TF-IDF-LR &
    \footnotesize $0.55$ & \footnotesize $0.57$ &
    \footnotesize{\color{gray} $0.48$} & \footnotesize{\color{gray}  $0.48$} &
    \footnotesize $0.50$ & \footnotesize $0.53$ \\
    
    TF-IDF-SVM &
    \cellcolor{blue!15}\footnotesize $\bf 0.58$ & \cellcolor{blue!15}\footnotesize $\bf 0.60$ &
    \footnotesize{\color{gray} $0.47$} & \footnotesize{\color{gray} $0.47$} &
    \cellcolor{blue!15}\footnotesize $\bf 0.56$ & \cellcolor{blue!15}\footnotesize $\bf 0.59$ \\

    \hline
    
    RoBERTa &
    \footnotesize $0.51$ & \footnotesize $0.55$ &
    \footnotesize{\color{gray} $0.47$} & \footnotesize{\color{gray} $0.48$} &
    \footnotesize $0.45$ & \footnotesize $0.51$ \\
    
    XLM-R &
    \footnotesize $0.49$ & \footnotesize $0.52$ &
    \footnotesize{\color{gray} $0.47$} & \footnotesize{\color{gray}  $0.48$} &
    \footnotesize $0.42$ & \footnotesize $0.47$ \\

    \hline

    GPT-ALL &
    \footnotesize $0.56$ & \footnotesize $0.62$ &
    \footnotesize{\color{gray} $0.45$} & \footnotesize{\color{gray}  $0.46$} &
    \cellcolor{blue!15}\footnotesize $\bf 0.56$ & \cellcolor{blue!15}\footnotesize $\bf 0.67$ \\

    \hline
    \hline

    \multicolumn{2}{|l}{\cellcolor{gray!10}\footnotesize\texttt{product-category}} & 
    \multicolumn{5}{r|}{\cellcolor{gray!10}\footnotesize\textit{$22$ classes}} \\
    
    \hline

    RANDOM &
    \footnotesize $0.01$ & \footnotesize $0.01$ &
    \footnotesize $0.20$ & \footnotesize $0.22$ &
    \footnotesize $0.01$ & \footnotesize $0.01$ \\
    
    MAJORITY &
    \footnotesize $0.01$ & \footnotesize $0.01$ &
    \footnotesize $0.24$ & \footnotesize $0.24$ &
    \footnotesize $0.00$ & \footnotesize $0.00$ \\
    
    \hline
    
    TF-IDF-KNN &
    \footnotesize $0.42$ & \footnotesize $0.47$ &
    \footnotesize $0.78$ & \footnotesize $0.79$ &
    \footnotesize $0.43$ & \footnotesize $0.49$ \\
    
    TF-IDF-LR &
    \footnotesize $0.52$ & \footnotesize $0.57$ &
    \cellcolor{blue!15}\footnotesize $\bf 0.87$ & \cellcolor{blue!15}\footnotesize $\bf 0.88$ &
    \footnotesize $0.49$ & \footnotesize $0.56$ \\
    
    TF-IDF-SVM &
    \footnotesize $0.59$ & \footnotesize $0.60$ &
    \footnotesize $0.86$ & \footnotesize $0.88$ &
    \footnotesize $0.59$ & \footnotesize $0.61$ \\
    
    \hline
    
    RoBERTa &
    \footnotesize $0.30$ & \footnotesize $0.53$ &
    \footnotesize $0.68$ & \footnotesize $0.90$ &
    \footnotesize $0.26$ & \footnotesize $0.48$ \\
    
    XLM-R &
    \footnotesize $0.35$ & \footnotesize $0.52$ &
    \footnotesize $0.78$ & \footnotesize $0.89$ &
    \footnotesize $0.30$ & \footnotesize $0.48$ \\

    \hline

    GPT-ALL &
    \cellcolor{blue!15}\footnotesize $\bf 0.69$ & \cellcolor{blue!15}\footnotesize $\bf 0.72$ &
    \footnotesize $0.84$ & \footnotesize $0.85$ &
    \cellcolor{blue!15}\footnotesize $\bf 0.73$ & \cellcolor{blue!15}\footnotesize $\bf 0.75$ \\

    \hline
    \hline

    \multicolumn{2}{|l}{\cellcolor{gray!10}\footnotesize\texttt{hazard}} & 
    \multicolumn{5}{r|}{\cellcolor{gray!10}\footnotesize\textit{$79$ / $261$ classes}} \\
    
    \hline

    RANDOM &
    \footnotesize $0.00$ & \footnotesize $0.00$ &
    \footnotesize $0.00$ & \footnotesize $0.00$ &
    \footnotesize $0.00$ & \footnotesize $0.00$ \\
    
    MAJORITY &
    \footnotesize $0.00$ & \footnotesize $0.00$ &
    \footnotesize $0.00$ & \footnotesize $0.00$ &
    \footnotesize $0.00$ & \footnotesize $0.00$ \\
    
    \hline
    
    TF-IDF-KNN &
    \footnotesize $0.28$ & \footnotesize $0.30$ &
    \footnotesize $0.71$ & \footnotesize $0.75$ &
    \footnotesize $0.28$ & \footnotesize $0.30$ \\
    
    TF-IDF-LR &
    \cellcolor{blue!15}\footnotesize $\bf 0.41$ & \cellcolor{blue!15}\footnotesize $\bf 0.44$ &
    \footnotesize $0.83$ & \footnotesize $0.85$ &
    \cellcolor{blue!15}\footnotesize $\bf 0.39$ & \cellcolor{blue!15}\footnotesize $\bf 0.43$ \\
    
    TF-IDF-SVM &
    \footnotesize $0.23$ & \footnotesize $0.24$ &
    \footnotesize $0.84$ & \footnotesize $0.86$ &
    \footnotesize $0.17$ & \footnotesize $0.18$ \\
    
    \hline
    
    RoBERTa &
    \footnotesize $0.36$ & \footnotesize $0.41$ &
    \cellcolor{blue!15}\footnotesize $\bf 0.85$ & \cellcolor{blue!15}\footnotesize $\bf 0.86$ &
    \footnotesize $0.32$ & \footnotesize $0.37$ \\
    
    XLM-R &
    \footnotesize $0.33$ & \footnotesize $0.38$ &
    \footnotesize $0.83$ & \footnotesize $0.86$ &
    \footnotesize $0.29$ & \footnotesize $0.35$ \\

    \hline
    \hline

    \multicolumn{2}{|l}{\cellcolor{gray!10}\footnotesize\texttt{product}} & 
    \multicolumn{5}{r|}{\cellcolor{gray!10}\footnotesize\textit{$184$ / $1256$ classes}} \\
    
    \hline

    RANDOM &
    \footnotesize $0.00$ & \footnotesize $0.00$ &
    \footnotesize $0.00$ & \footnotesize $0.00$ &
    \footnotesize $0.00$ & \footnotesize $0.00$ \\
    
    MAJORITY &
    \footnotesize $0.00$ & \footnotesize $0.00$ &
    \footnotesize $0.00$ & \footnotesize $0.00$ &
    \footnotesize $0.00$ & \footnotesize $0.00$ \\
    
    \hline
    
    TF-IDF-KNN &
    \footnotesize $0.32$ & \footnotesize $0.35$ &
    \footnotesize $0.46$ & \footnotesize $0.51$ &
    \footnotesize $0.24$ & \footnotesize $0.35$ \\
    
    TF-IDF-LR &
    \cellcolor{blue!15}\footnotesize $\bf 0.51$ & \cellcolor{blue!15}\footnotesize $\bf 0.53$ &
    \cellcolor{blue!15}\footnotesize $\bf 0.68$ & \cellcolor{blue!15}\footnotesize $\bf 0.72$ &
    \cellcolor{blue!15}\footnotesize $\bf 0.43$ & \cellcolor{blue!15}\footnotesize $\bf 0.52$ \\
    
    TF-IDF-SVM &
    \footnotesize $0.09$ & \footnotesize $0.10$ &
    \footnotesize $0.50$ & \footnotesize $0.52$ &
    \footnotesize $0.00$ & \footnotesize $0.00$ \\
    
    \hline
    
    RoBERTa &
    \footnotesize $0.43$ & \footnotesize $0.57$ &
    \footnotesize $0.56$ & \footnotesize $0.72$ &
    \footnotesize $0.28$ & \footnotesize $0.52$ \\
    
    XLM-R &
    \footnotesize $0.28$ & \footnotesize $0.54$ &
    \footnotesize $0.44$ & \footnotesize $0.76$ &
    \footnotesize $0.14$ & \footnotesize $0.41$ \\
    
    \hline

\end{tabular}
}
\caption{Macro-averaged $F_1$-score (\textit{mean \& maximum}) over 5 CV splits. Blue cells highlight the best mean score per column and classification task. For \texttt{hazard} and \texttt{product} scores are only evaluated on well-supported classes. The grayed scores are added for completeness, but not comparable, as $\mathcal{C}_{high}$ of \texttt{hazard-category} consists of only one class.}
\label{tab:results}
\end{center}
\end{table}

\vspace{-5pt}
\subsubsection{Traditional Baselines}
The $F_1$-scores on classification (overall, as well as $\mathcal{C}_{high}$ and $\mathcal{C}_{low}$ introduced in Figure~\ref{fig:label_distribution}) are presented in Table~\ref{tab:results}. All the classifiers outperform both naive baselines, except for the $\mathcal{C}_{high}$ of \texttt{hazard-category}. In this task, $\mathcal{C}_{high}$ only consists of the majority class. This means that classification on this task is not useful as there is only one label present in the test set. Every class not predicted to this label (even if correctly predicted in general) will therefore diminish the macro $F_1$.
Since the BOW-based classifiers in general perform worse than TF-IDF, we only report them in the Appendix (see Table~\ref{tab:results_extended}).

The best-performing non-prompting classifier for the \texttt{hazard-category} and \texttt{product-category} tasks is TF-IDF-SVM as it excels on $\mathcal{C}_{low}$. For the fine-grained \texttt{hazard} and \texttt{product} tasks the SVM's performance on the less represented classes drops, and TF-IDF-LR takes the position as the best traditional classifier. In parallel, TF-IDF-SVM's relative $F_1$-score increases on high support classes, which we interpret as an inability to learn from classes with very few samples, paired with good performance on training sample counts $> 100$.
TF-IDF-KNN has the overall lowest performance of the traditional classifiers.

\vspace{-5pt}
\subsubsection{Transformers}
TF-IDF-LR performs overall comparably or better than the best encoder-only Transformer RoBERTa. We assume that the Transformer architectures would need higher numbers of training samples per class to perform on par or better compared to traditional methods.

Interestingly, RoBERTa and XLM-R both achieve lower and less stable $F_1$-score for \texttt{product} and \texttt{product-category}. This is not seen in other classifiers. By inspecting the data, we see that in both of these tasks brand names (e.g. ``SnoBalls'', ``Sriracha'') or hard-to-define word creations (e.g. ``pecan caramel stars'', ``Vanilla Cups'') make up the relevant part for classification. These are hard to learn during fine-tuning, as they are both very specific and most probably underrepresented in the data used for pre-training.
Analysis of the spans in \texttt{product-title} shows that TF-IDF-LR often infers the product from the supplier and/or the hazard in such cases. 

Surprisingly, the multilingual XLM-R only outperforms RoBERTa in the $\mathcal{C}_{high}$ of the \texttt{hazard-category} and \texttt{product-category} tasks even though the texts come in multiple languages. We assume the low number of per-class samples is not sufficient for the very large embedding layer of XLM-R.
Analysis of the performance on the two most common languages, English and German, is shown in Appendix~\ref{sec:appendix_results}, Table~\ref{tab:langauge_based_assessment}. The results suggest that while XLM-R performs closer to RoBERTa on non-English texts, its performance in terms of $F_1$ is generally worse than that of RoBERTa. XLM-R performs especially worse than RoBERTa on German texts in tasks with low per-class support (namely \texttt{hazard} and \texttt{product}), which supports our claim that XLM-R needs more training examples than RoBERTa to deliver good performance.

\subsubsection{Large Language Models}

A simple prompting approach (GPT-ALL) performs comparably to TF-IDF-LR and RoBERTa on the \texttt{hazard-category} and is the best classifier for \texttt{product-category}. In both cases, it excels on the $\mathcal{C}_{low}$ as it only needs two labeled samples per class. Since when prompting, the predicted class is delivered in free text, the LLM may produce output that is not within the set of class labels. While these outputs could be interpreted as belonging to one of the class labels, we only count exact matches. In both coarse tasks, GPT-ALL failed to predict any class for $2\%$ of the samples (Table~\ref{tab:results_prompting}).
In a qualitative analysis, shown in Appendix~\ref{sec:appendix_results}, we see that while some of the prompting failures can be improved by better prompt design (e.g. more samples per class, see also Table~\ref{tab:results_prompting}), others can even be seen as a positive feature of prompting.

\begin{table}[!hb]
\begin{center}
\resizebox*{\linewidth}{!}{
\begin{tabular}{|r|ccc|ccc|}

    \hline
    
    \multicolumn{1}{|c|}{\cellcolor{gray!25}\textbf{Model}} &
    \multicolumn{1}{m{1cm}<{\centering} }{\cellcolor{gray!25} \footnotesize $\bf F_1$ (\textit{macro})} &
    \multicolumn{1}{m{1cm}<{\centering} }{\cellcolor{gray!25} \footnotesize \textbf{LLM usage}} &
    \multicolumn{1}{m{1cm}<{\centering}|}{\cellcolor{gray!25} \footnotesize \textbf{Prompt Size} (\textit{chars})}  &
    \multicolumn{1}{m{1cm}<{\centering} }{\cellcolor{gray!25} \footnotesize \textbf{Classes per Prompt}} &
    \multicolumn{1}{m{1cm}<{\centering} }{\cellcolor{gray!25} \footnotesize \textbf{Samples per Class}} &
    \multicolumn{1}{m{1cm}<{\centering}|}{\cellcolor{gray!25} \footnotesize \textbf{Failure Rate}}\\

    \hline
    \hline

    \multicolumn{2}{|l}{\cellcolor{gray!10}\footnotesize\texttt{hazard-category}} & 
    \multicolumn{5}{r|}{\cellcolor{gray!10}\footnotesize\textit{$10$ classes; $\alpha=0.05$}} \\

    \hline
    
    TF-IDF-LR &
    \footnotesize $0.55$ &
    \footnotesize - &
    \footnotesize - &
    \footnotesize - &
    \footnotesize - &
    \footnotesize - \\
    
    GPT-ALL &
    \footnotesize $0.53$ &
    \footnotesize $100\%$ &
    \footnotesize $2301.7$ &
    \footnotesize $10.0$ &
    \footnotesize $2.0$ &
    \footnotesize $2\%$ \\

    \hline

    GPT-SIM-10 &
    \footnotesize $0.52$ &
    \footnotesize $100\%$ &
    \footnotesize $1206.3$ &
    \footnotesize $\enspace 5.8$ &
    \footnotesize $1.7$ &
    \footnotesize $2\%$ \\

    GPT-MAX-5 &
    \footnotesize $0.55$ &
    \footnotesize $100\%$ &
    \footnotesize $1214.9$ &
    \footnotesize $\enspace 5.0$ &
    \footnotesize $2.0$ &
    \footnotesize $2\%$ \\

    GPT-SIM-20 &
    \footnotesize $0.53$ &
    \footnotesize $100\%$ &
    \footnotesize $2301.7$ &
    \footnotesize $10.0$ &
    \footnotesize $2.0$ &
    \footnotesize $2\%$ \\

    GPT-MAX-10 &
    \footnotesize $0.49$ &
    \footnotesize $100\%$ &
    \footnotesize $2301.7$ &
    \footnotesize $10.0$ &
    \footnotesize $2.0$ &
    \footnotesize $3\%$ \\

    GPT-CICLe &
    \cellcolor{blue!15}\footnotesize $\bf 0.56$ &
    \cellcolor{green!15}\footnotesize $\bf\enspace 60\%$ &
    \cellcolor{green!15}\footnotesize $\bf\enspace 707.5$ &
    \footnotesize $\enspace 2.6$ &
    \footnotesize $2.0$ &
    \footnotesize $2\%$ \\

    \hline
    \hline

    \multicolumn{2}{|l}{\cellcolor{gray!10}\footnotesize\texttt{product-category}} & 
    \multicolumn{5}{r|}{\cellcolor{gray!10}\footnotesize\textit{$22$ classes; $\alpha=0.05$}} \\

    \hline

    TF-IDF-LR &
    \footnotesize $0.52$ &
    \footnotesize - &
    \footnotesize - &
    \footnotesize - &
    \footnotesize - &
    \footnotesize - \\

    GPT-ALL &
    \footnotesize $0.72$ &
    \footnotesize $100\%$ &
    \footnotesize $5621.1$ &
    \footnotesize $22.0$ &
    \footnotesize $2.0$ &
    \footnotesize $2\%$ \\

    \hline

    GPT-SIM-10 &
    \footnotesize $0.66$ &
    \footnotesize $100\%$ &
    \cellcolor{green!15}\footnotesize $\bf 1378.6$ &
    \footnotesize $\enspace 6.6$ &
    \footnotesize $1.5$ &
    \footnotesize $4\%$ \\

    GPT-MAX-5 &
    \footnotesize $0.64$ &
    \footnotesize $100\%$ &
    \footnotesize $1395.6$ &
    \footnotesize $\enspace 5.0$ &
    \footnotesize $2.0$ &
    \footnotesize $4\%$ \\

    GPT-SIM-20 &
    \footnotesize $0.69$ &
    \footnotesize $100\%$ &
    \footnotesize $2614.0$ &
    \footnotesize $11.6$ &
    \footnotesize $1.7$ &
    \footnotesize $2\%$ \\

    GPT-MAX-10 &
    \footnotesize $0.68$ &
    \footnotesize $100\%$ &
    \footnotesize $2631.8$ &
    \footnotesize $10.0$ &
    \footnotesize $2.0$ &
    \footnotesize $6\%$ \\

    GPT-CICLe &
    \cellcolor{blue!15}\footnotesize $\bf 0.69$ &
    \cellcolor{green!15}\footnotesize $\bf\enspace 98\%$ &
    \footnotesize $1620.1$ &
    \footnotesize $\enspace 5.9$ &
    \footnotesize $2.0$ &
    \footnotesize $5\%$ \\

    \hline
    \hline

    \multicolumn{2}{|l}{\cellcolor{gray!10}\footnotesize\texttt{hazard}} & 
    \multicolumn{5}{r|}{\cellcolor{gray!10}\footnotesize\textit{$102$ / $261$ classes; $\alpha=0.20$}} \\

    \hline

    TF-IDF-LR &
    \footnotesize $0.37$ &
    \footnotesize - &
    \footnotesize - &
    \footnotesize - &
    \footnotesize - &
    \footnotesize - \\
    
    \hline
    
    GPT-SIM-10 &
    \cellcolor{blue!15}\footnotesize $\bf 0.44$ &
    \footnotesize $100\%$ &
    \footnotesize $1308.0$ &
    \footnotesize $\enspace 7.5$ &
    \footnotesize $1.3$ &
    \footnotesize $15\%$ \\

    GPT-MAX-5 &
    \footnotesize $0.42$ &
    \footnotesize $100\%$ &
    \footnotesize $1335.4$ &
    \footnotesize $\enspace 5.0$ &
    \footnotesize $2.0$ &
    \footnotesize $14\%$ \\

    GPT-SIM-20 &
    \cellcolor{blue!15}\footnotesize $\bf 0.44$ &
    \footnotesize $100\%$ &
    \footnotesize $2457.0$ &
    \footnotesize $14.2$ &
    \footnotesize $1.4$ &
    \footnotesize $11\%$ \\

    GPT-MAX-10 &
    \footnotesize $0.42$ &
    \footnotesize $100\%$ &
    \footnotesize $2521.1$ &
    \footnotesize $10.0$ &
    \footnotesize $2.0$ &
    \footnotesize $13\%$ \\

    GPT-CICLe &
    \footnotesize $0.42$ &
    \cellcolor{green!15}\footnotesize $\bf\enspace 87\%$ &
    \cellcolor{green!15}\footnotesize $\bf 1251.6$ &
    \footnotesize $\enspace 4.7$ &
    \footnotesize $2.0$ &
    \footnotesize $14\%$ \\

    \hline
    \hline

    \multicolumn{2}{|l}{\cellcolor{gray!10}\footnotesize\texttt{product}} & 
    \multicolumn{5}{r|}{\cellcolor{gray!10}\footnotesize\textit{$257$ / $1256$ classes; $\alpha=0.40$}} \\

    \hline

    TF-IDF-LR &
    \footnotesize $0.54$ &
    \footnotesize - &
    \footnotesize - &
    \footnotesize - &
    \footnotesize - &
    \footnotesize - \\

    \hline

    GPT-SIM-10 &
    \footnotesize $0.54$ &
    \footnotesize $100\%$ &
    \cellcolor{green!15}\footnotesize $\bf 1238.7$ &
    \footnotesize $\enspace 8.5$ &
    \footnotesize $1.2$ &
    \footnotesize $34\%$ \\

    GPT-MAX-5 &
    \footnotesize $0.56$ &
    \footnotesize $100\%$ &
    \footnotesize $1270.0$ &
    \footnotesize $\enspace 5.0$ &
    \footnotesize $2.0$ &
    \footnotesize $20\%$ \\

    GPT-SIM-20 &
    \footnotesize $0.53$ &
    \footnotesize $100\%$ &
    \footnotesize $2324.3$ &
    \footnotesize $16.7$ &
    \footnotesize $1.2$ &
    \footnotesize $36\%$ \\

    GPT-MAX-10 &
    \footnotesize $0.57$ &
    \footnotesize $100\%$ &
    \footnotesize $2385.2$ &
    \footnotesize $10.0$ &
    \footnotesize $2.0$ &
    \footnotesize $23\%$ \\

    GPT-CICLe &
    \cellcolor{blue!15}\footnotesize $\bf 0.58$ &
    \cellcolor{green!15}\footnotesize $\bf\enspace 96\%$ &
    \footnotesize $1645.8$ &
    \footnotesize $\enspace 6.7$ &
    \footnotesize $2.0$ &
    \footnotesize $24\%$ \\

    \hline
\end{tabular}
}
\caption{Comparison of reduced prompting strategies. Blue cells highlight the best performance of reduced prompts per column and classification task; green cells the most resource-friendly value. For \texttt{hazard} and \texttt{product} $F_1$ is only evaluated on well-supported classes.}
\label{tab:results_prompting}\vspace{-20pt}
\end{center}
\end{table}

\begin{figure*}[t]
\begin{center}
\includegraphics[height=4cm, trim={0.0cm .1cm .3cm .1cm}, clip]{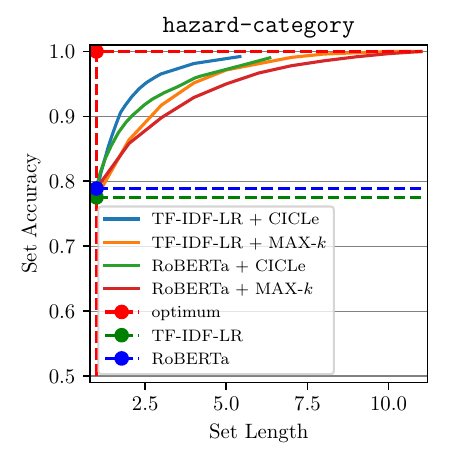}
\includegraphics[height=4cm, trim={1.5cm .1cm .3cm .1cm}, clip]{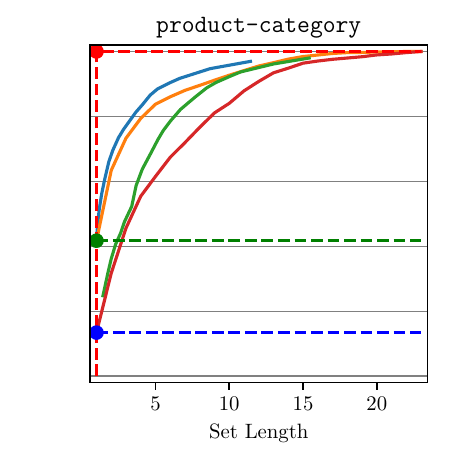}
\includegraphics[height=4cm, trim={1.5cm .1cm .3cm .1cm}, clip]{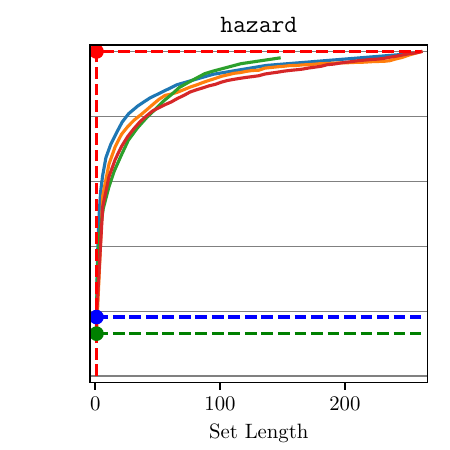}
\includegraphics[height=4cm, trim={1.5cm .1cm .3cm .1cm}, clip]{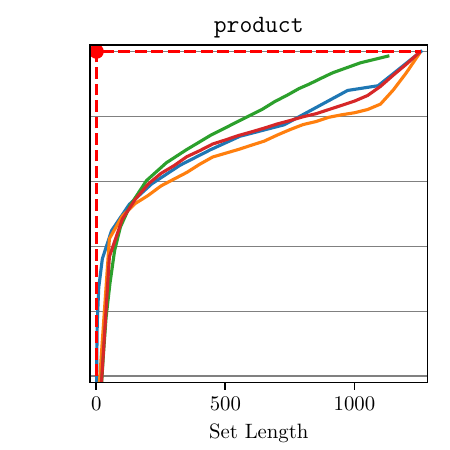}
\caption{CP performance on the best traditional classifier (green dashed line) and the best encoder-only Transformer (blue dashed line). The optimal prediction set (red dashed line) would have one element, which is the true class with a probability of 1.}
\label{fig:conformal_performance}\vspace{-10pt}
\end{center}
\end{figure*}

\paragraph{Reduced prompts} Naive few-shot prompting (e.g., with two shots per class) may work for few but is not optimal for a high number of classes. In order to reduce the number of classes for which we include shots in the prompt, we compare three approaches that meaningfully reduce the context length. A condensed version of the results for reduced prompting is shown in Table~\ref{tab:results_prompting} (the full version is in Table~\ref{tab:results_prompting_extended} of the Appendix~\ref{sec:appendix_results}).
Simply reducing the number of few-shot samples by moving to the GPT-SIM-$k$ prompts also reduces the overall $F_1$-scores\footnote{GPT-ALL and GPT-SIM-20 are identical in \texttt{hazard-category}} as performance on $\mathcal{C}_{low}$ drops (see Table~\ref{tab:results_prompting_extended}). This is largely owed to the circumstance that the $\mathcal{C}_{low}$ makes up the bulk of the classes and we are more likely to remove samples of a low-support class than the less prominent high-support classes. The latter actually profit from the reduction, as they now compete with fewer other classes. An analysis of the effect of reducing classes in the prompts on maximum and minimum performance is shown in Appendix~\ref{sec:appendix_results}.
Nevertheless, this very simple setup is already very powerful and outperforms the more sophisticated prompts on \texttt{product}.

A noteworthy detail of the \texttt{hazard-category} task is that it contains exactly ten classes. This means that GPT-ALL, GPT-SIM-20, and GPT-MAX-10 are identical apart from sample order: while GPT-ALL and GPT-SIM-20 are ordered by cosine similarity, GPT-MAX-10 is ordered by class probability as determined by the base classifier. While the evidence is anecdotal, the $F_1$-scores suggest that sample order is important in prompting and that ordering by similarity may be better.

An estimate of the capabilities of *-MAX-$k$ vs. *-CICLe is shown in Figure~\ref{fig:conformal_performance}: CP leads to more concise prediction sets (i.e. smaller sets at the same set accuracy) than GPT-MAX-$k$. In contrast to max-$k$, CP produces shorter sets if the classifier has a high certainty on the true class. Nevertheless, classifiers with lower overall performance will improve on max-$k$'s accuracy only at larger sets.
This is also reflected in Table~\ref{tab:results_prompting}: GPT-CICLe performs better or at least equal to GPT-MAX-$k$ in all tasks.

\noindent\textbf{Focusing on GPT-CICLe}, Table~\ref{tab:results_prompting} shows that it outperforms its base classifier TF-IDF-LR on all tasks as well as all reduced prompting baselines for three out of four tasks.
For higher numbers of classes, the performance of GPT-CICLe drops in $\mathcal{C}_{low}$ compared to the other prompts (see Table~\ref{tab:results_prompting_extended}). We attribute this to the higher $\alpha$ in these cases: here we sacrifice CP guarantees for faster prediction.
GPT-CICLe also produces the lowest prompt lengths at the same $F_1$ (Table~\ref{tab:results_prompting}). Here, we argue, that the base classifier used by GPT-MAX-$k$ and GPT-CICLe is able to produce more concise contexts for the LLM than simple cosine similarity. While this is already a step towards energy-efficient prompting, CICLe can avoid accessing the LLM completely in some cases: the extent depends on the task and the  $\alpha$, but ranges between accessing the LLM in only $60\%$ to $98\%$ of cases in our tasks.
The selection of alpha can in theory be done through an extensive hyperparameter search. As this becomes very computationally expensive for LLMs, we select $\alpha$ based on Figure~\ref{fig:conformal_performance} in our experiments (see Appendix~\ref{sec:appendix_models}). We leave computationally efficient optimization of $\alpha$ for future research.

\vspace{-8pt}
\section{Discussion}

\vspace{-8pt}
Our results suggest that a general reduction of few-shot examples, taken from a population that is very likely to contain the true class, can outperform traditional ML classifiers and Transformers in terms of $F_1$ even without using other prompting techniques, e.g., example matching~\cite{Ahmed2023_FewShot+StaticAnalysis} or CoT.
This makes simple selection of few shots based on cosine similarity (GPT-SIM-$k$) a powerful approach, especially for tasks with a high number of classes or extraction tasks. In this setting, we leverage the well-documented strong in-context reasoning capabilities of LLMs \cite{Chowdhery_palm, Wei2022_CoT, Shi2023_PromptSpace}, which do not necessarily need a full view of all possible classes.
However, in cases where we are able to train a decently performing base classifier, CICLe is a better choice: it boosts the performance of the base classifier by detecting insecure samples using CP and redirecting them to an LLM. Further analysis (see Appendix~\ref{sec:appendix_cm}) shows that CICLe's strength is mainly based on better performance on low-support classes compared to the base classifier. This is most probably owed to the low demand for ``training data'' in few-shot prompting. In general, the influence of the base classifier and LLM on CICLe's performance is controlled by the $\alpha$ value: low values favor the LLM, while high values favor the base classifier.
Lastly, CICLe is subject to a trade-off between predictive and temporal performance: it allows sacrificing accuracy (e.g. due to hardware and time constraints) while keeping the context for the LLM as concise as possible. This makes it a resource-friendly alternative to prompting with all classes.
Our results further suggest, in line with \citet{liu_contexts}, that sample order and context length are highly influential on prompting performance. We therefore see a need for future work on optimal ordering of few-shot samples and tuning of the $\alpha$ parameter within CICLe.  

\vspace{-8pt}
\section{Conclusions}\label{sec:conclusion}

\vspace{-8pt}
We present the first dataset for large-scale multi-class classification of short texts describing food recalls, which we release publicly under a \href{https://creativecommons.org/licenses/by-nc-sa/4.0/deed.en}{CC BY-NC-SA~4.0} license. The dataset contains labels annotated by experts on food products and hazards on two levels of granularity, coarse (tens) and fine (hundreds).

Answering our research question \textbf{R1}, we find that the titles of food recall announcements mostly hold sufficient information for classification, although not all of them include information on both product and hazards (\textbf{R1.1}).
A strong indication for this statement is that we are able to reach above random performance with multiple classifiers: we carry out benchmarking, analysing the performance of naive, traditional, and deep-learning classifiers for all four classification tasks in the proposed dataset, showing the potential of traditional classifiers on this dataset. Our results further indicate that products are harder to classify than hazards, which we attribute to the often creative choice of product names (\textbf{R1.2}).

Moving on to \textbf{R2}, we also experimented with prompting using one few-shot example per class. Here we see that gpt-3.5 performs comparably or better than the best non-LLM classifiers (\textbf{R2.1}). We also show that resource-efficient prompting at comparable predictive performance is possible by proposing a novel LLM-in-the-loop framework that leverages Conformal Prediction (\textbf{R2.2}).

\section{Limitations}\label{sec:limitations}
Nevertheless, the dataset and approach discussed in this paper are subject to limitations. Regarding the dataset, we identified the following:
\begin{itemize}
    \item The labels in our dataset are subject to noise. Specifically, some samples are missing evidence for one or more of the assigned classes, while tokens important for classification may indicate more than one class. This may lead to classifiers trained on the data seeing contradicting examples and therefore limit their predictive performance. 

    \item The spans in \texttt{hazard-title} and \texttt{product-title} are machine generated and not manually curated. This means that while they give an estimation of word importance, they are no gold standard for explainability tasks.
\end{itemize}
We aim to improve these limitations in future iterations of the dataset. For our approach leveraging CP for few-shot prompting, we found the following limitations:
\begin{itemize}
    \item As visualized in Figure~\ref{fig:conformal_performance}, CP represents a trade-off between high prediction set accuracy and low set length relative to the total number of classes. This means that with a rising total number of classes, we will have to sacrifice predictive performance in order to keep the prompt size feasible, which ultimately might render the approach useless.

    \item In this paper we used normal conformal prediction, which only guarantees a certain probability of a single true class being in the prediction set. In order to achieve true multilabel guarantees we would need to switch to mondrian CP.

    \item We only verify our approach on a single LLM. This means that our approach might not be generalizable to other LLMs and perform differently or not at all for few-shot prompting with such models.

\end{itemize}
While we have to accept the first of these points as inherent to the approach, we are planning to address the remaining points in future work. As all the data used in our dataset was already publicly available before the publication of this work, we do not violate anybody's privacy by republishing it.

\section*{Acknowledgements}

\anonymous[This work was funded by the European Union.]{Funding for this research has been provided by the European Union’s Horizon Europe research and innovation programme EFRA (Grant Agreement Number 101093026). Funded by the European Union. Views and opinions expressed are however those of the author(s) only and do not necessarily reflect those of the European Union or European Commission-EU. Neither the European Union nor the granting authority can be held responsible for them.} {\normalsize\euflag} 

This work has been partially supported by project
MIS 5154714 of the National Recovery and Resilience Plan Greece 2.0 funded by the European
Union under the NextGenerationEU Program.

\raggedbottom
\bibliography{anthology,custom}

\pagebreak

\appendix
\section*{Appendix}

\FloatBarrier

\section{Data sample and statistics:}\label{sec:appendix_data}
\begin{figure*}[ht]
    \centering
    \includegraphics[width=.8\linewidth, trim={1cm 1cm 0cm 0cm}, clip]{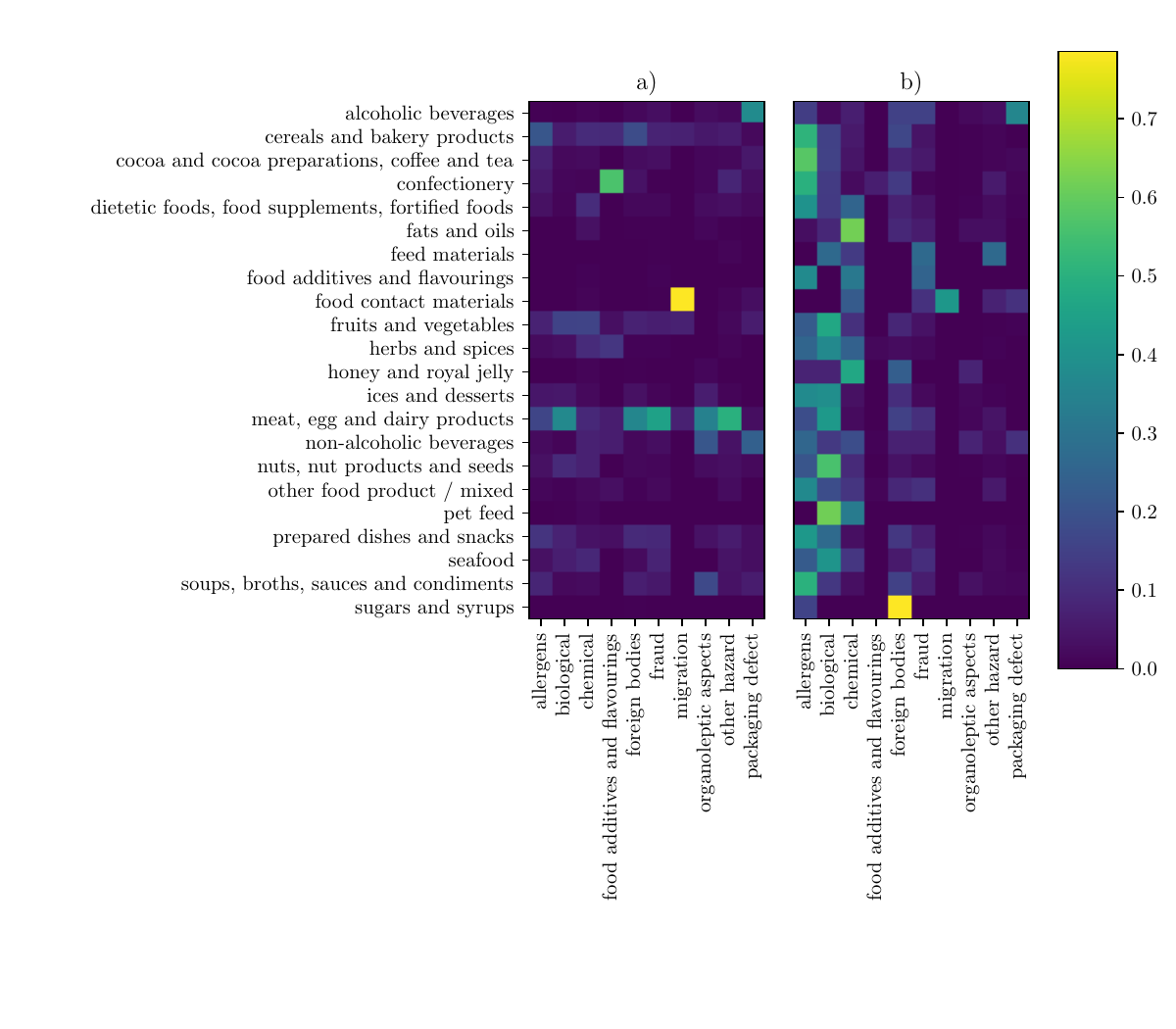}
    \caption{Label co-occurence for the \texttt{hazard-category} and \texttt{product-category} tasks normalized by \textbf{a)}~\texttt{hazard-category}, and \textbf{b)}~\texttt{product-category}. While this is not true for all the \texttt{hazard-category}-\texttt{product-category} pairs, some show strong linkage.}
    \label{fig:label_co-occurence}
\end{figure*}

In this appendix, we provide additional insights into the dataset shared with this publication. Table~\ref{tab:domains} lists the original websites our texts were collected from, Table~\ref{tab:data_statistics} provides a statistical overview over the class- and language-distribution in the data, 
and Figure~\ref{fig:label_co-occurence} is a heatmap highlighting co-occurrence of classes in the \texttt{hazard-category} and \texttt{product-category} tasks. Finally, we share a few data samples along with their labels in Figure~\ref{fig:example_texts}.

\begin{table}[hb]
\begin{center}
\begin{tabular}{|l|r|}

    \hline
    \cellcolor{gray!25}\textbf{Domain}                  & \cellcolor{gray!25}\textbf{Samples}     \\
    \hline
    \hline
    {\footnotesize www.fda.gov:                        } & {\footnotesize 1740} \\
    {\footnotesize www.fsis.usda.gov:                  } & {\footnotesize 1112} \\
    {\footnotesize www.productsafety.gov.au:           } & {\footnotesize  925} \\
    {\footnotesize www.food.gov.uk:                    } & {\footnotesize  902} \\
    {\footnotesize www.lebensmittelwarnung.de:         } & {\footnotesize  886} \\
    {\footnotesize www.inspection.gc.ca:               } & {\footnotesize  864} \\
    {\footnotesize www.fsai.ie:                        } & {\footnotesize  358} \\
    {\footnotesize www.foodstandards.gov.au:           } & {\footnotesize  281} \\
    {\footnotesize inspection.canada.ca:               } & {\footnotesize  124} \\
    {\footnotesize www.cfs.gov.hk:                     } & {\footnotesize  123} \\
    {\footnotesize recalls-rappels.canada.ca:          } & {\footnotesize   96} \\
    {\footnotesize tna.europarchive.org:               } & {\footnotesize   52} \\
    {\footnotesize wayback.archive-it.org:             } & {\footnotesize   23} \\
    {\footnotesize healthycanadians.gc.ca:             } & {\footnotesize   18} \\
    {\footnotesize www.sfa.gov.sg:                     } & {\footnotesize   11} \\
    {\footnotesize www.collectionscanada.gc.ca:        } & {\footnotesize   10} \\
    {\footnotesize securite-alimentaire.public.lu:     } & {\footnotesize    8} \\
    {\footnotesize portal.efet.gr:                     } & {\footnotesize    4} \\
    {\footnotesize www.foodstandards.gov.scot:         } & {\footnotesize    3} \\
    {\footnotesize www.ages.at:                        } & {\footnotesize    2} \\
    {\footnotesize www.accessdata.fda.gov:             } & {\footnotesize    1} \\
    {\footnotesize webarchive.nationalarchives.gov.uk: } & {\footnotesize    1} \\
    {\footnotesize www.salute.gov.it:                  } & {\footnotesize    1} \\
    {\footnotesize www.foedevarestyrelsen.dk:          } & {\footnotesize    1} \\
    \hline

\end{tabular}
\caption{Data sources, ordered by support number}
\label{tab:domains}\vspace{-15pt}
\end{center}
\end{table}

\begin{table*}[ht]
    \centering
    \resizebox*{!}{.8\textheight}{
        \setlength{\tabcolsep}{4pt}
        \begin{tabular}{|cr||rl|rl||rl|rl|rl||rl|}
            \cline{3-14}

            \multicolumn{2}{c|}{} &
            \multicolumn{4}{c||}{\cellcolor{gray!25}\textbf{Per Class}} &
            \multicolumn{6}{c||}{\cellcolor{gray!25}\textbf{Support Based}} &
            \multicolumn{2}{c|}{\cellcolor{gray!25}} \\

            \multicolumn{2}{c|}{} &
            \multicolumn{2}{c|}{\cellcolor{gray!25}\footnotesize \textit{most freq.}} &
            \multicolumn{2}{c||}{\cellcolor{gray!25}\footnotesize \textit{least freq.}} &
            \multicolumn{2}{c|}{\cellcolor{gray!25}\footnotesize $\mathcal{C}_{high}$} &
            \multicolumn{2}{c|}{\cellcolor{gray!25}\footnotesize $\mathcal{C}_{medium}$} &
            \multicolumn{2}{c||}{\cellcolor{gray!25}\footnotesize $\mathcal{C}_{low}$} &
            \multicolumn{2}{c|}{\cellcolor{gray!25}\multirow{-2}{*}{\textbf{Total}}} \\

            \cline{3-14}

            \multicolumn{2}{l}{\texttt{hazard-category}} &
            \multicolumn{2}{c}{\tiny{biological}} &
            \multicolumn{2}{c}{\tiny{migration}} &
            \multicolumn{8}{c}{} \\
            
            \hline
            
            \cellcolor{gray!25} & \cellcolor{gray!25} 1994 - 1998 & $34$ & $(33.6)$ &  &  & $34$ & $(33.6)$ & $4$ & $(71.0)$ & $20$ & $(34.6)$ & \cellcolor{gray!10}$58$ & \cellcolor{gray!10}$(36.5)$\\
            \cellcolor{gray!25} & \cellcolor{gray!25} 1999 - 2002 & $50$ & $(50.8)$ &  &  & $50$ & $(50.8)$ & $21$ & $(52.9)$ & $59$ & $(46.0)$ & \cellcolor{gray!10}$130$ & \cellcolor{gray!10}$(49.0)$\\
            \cellcolor{gray!25} & \cellcolor{gray!25} 2003 - 2006 & $53$ & $(52.8)$ &  &  & $53$ & $(52.8)$ & $61$ & $(60.5)$ & $78$ & $(58.1)$ & \cellcolor{gray!10}$192$ & \cellcolor{gray!10}$(57.4)$\\
            \cellcolor{gray!25} & \cellcolor{gray!25} 2007 - 2010 & $158$ & $(95.3)$ &  &  & $158$ & $(95.3)$ & $49$ & $(81.0)$ & $112$ & $(72.5)$ & \cellcolor{gray!10}$319$ & \cellcolor{gray!10}$(85.1)$\\
            \cellcolor{gray!25} & \cellcolor{gray!25} 2011 - 2014 & $308$ & $(91.6)$ &  &  & $308$ & $(91.6)$ & $293$ & $(81.2)$ & $233$ & $(71.9)$ & \cellcolor{gray!10}$834$ & \cellcolor{gray!10}$(82.4)$\\
            \cellcolor{gray!25} & \cellcolor{gray!25} 2015 - 2018 & $911$ & $(91.5)$ &  &  & $911$ & $(91.5)$ & $1048$ & $(86.8)$ & $726$ & $(82.2)$ & \cellcolor{gray!10}$2685$ & \cellcolor{gray!10}$(87.1)$\\
            \cellcolor{gray!25} \multirow{-7}*{\rotatebox{90}{\textbf{By Year}}} & \cellcolor{gray!25} 2019 - 2022 & $1043$ & $(85.7)$ & $14$ & $(44.1)$ & $1043$ & $(85.7)$ & $1051$ & $(89.0)$ & $1234$ & $(83.7)$ & \cellcolor{gray!10}$3328$ & \cellcolor{gray!10}$(86.0)$\\
            
            \hline
            
            \cellcolor{gray!25} & \cellcolor{gray!25} DE & $278$ & $(58.8)$ & $10$ & $(29.0)$ & $278$ & $(58.8)$ & $96$ & $(52.7)$ & $514$ & $(58.4)$ & \cellcolor{gray!10}$888$ & \cellcolor{gray!10}$(57.9)$\\
            \cellcolor{gray!25} & \cellcolor{gray!25} DK &  &  &  &  &  &  & $1$ & $(42.0)$ &  &  & \cellcolor{gray!10}$1$ & \cellcolor{gray!10}$(42.0)$\\
            \cellcolor{gray!25} & \cellcolor{gray!25} EN & $2278$ & $(90.5)$ & $4$ & $(81.8)$ & $2278$ & $(90.5)$ & $2426$ & $(87.4)$ & $1940$ & $(85.2)$ & \cellcolor{gray!10}$6644$ & \cellcolor{gray!10}$(87.8)$\\
            \cellcolor{gray!25} & \cellcolor{gray!25} FR &  &  &  &  &  &  & $1$ & $(73.0)$ & $7$ & $(71.7)$ & \cellcolor{gray!10}$8$ & \cellcolor{gray!10}$(71.9)$\\
            \cellcolor{gray!25} & \cellcolor{gray!25} GR &  &  &  &  &  &  & $3$ & $(74.0)$ & $1$ & $(47.0)$ & \cellcolor{gray!10}$4$ & \cellcolor{gray!10}$(67.2)$\\
            \cellcolor{gray!25} \multirow{-6}*{\rotatebox{90}{\textbf{By Language}}} & \cellcolor{gray!25} IT & $1$ & $(51.0)$ &  &  & $1$ & $(51.0)$ &  &  &  &  & \cellcolor{gray!10}$1$ & \cellcolor{gray!10}$(51.0)$\\
            
            \hline
            \hline
            
            \multicolumn{2}{|c||}{\cellcolor{gray!25} \textbf{Total}} & \cellcolor{gray!10}$2557$ & \cellcolor{gray!10}$(87.0)$ & \cellcolor{gray!10}$14$ & \cellcolor{gray!10}$(44.1)$ & \cellcolor{gray!10}$2557$ & \cellcolor{gray!10}$(87.0)$ & \cellcolor{gray!10}$2527$ & \cellcolor{gray!10}$(86.0)$ & \cellcolor{gray!10}$2462$ & \cellcolor{gray!10}$(79.5)$ & \cellcolor{gray!10}$7546$ & \cellcolor{gray!10}$(84.2)$\\

            \hline

            \multicolumn{2}{l}{\texttt{product-category}} &
            \multicolumn{2}{c}{\tiny{meat, egg and dairy products}} &
            \multicolumn{2}{c}{\tiny{sugars and syrups}} &
            \multicolumn{8}{c}{} \\
            
            \hline
            
            \cellcolor{gray!25} & \cellcolor{gray!25} 1994 - 1998 & $43$ & $(32.6)$ &  &  & $48$ & $(34.9)$ & $5$ & $(46.2)$ & $5$ & $(41.8)$ & \cellcolor{gray!10}$58$ & \cellcolor{gray!10}$(36.5)$\\
            \cellcolor{gray!25} & \cellcolor{gray!25} 1999 - 2002 & $37$ & $(48.6)$ &  &  & $58$ & $(47.6)$ & $29$ & $(47.5)$ & $43$ & $(51.8)$ & \cellcolor{gray!10}$130$ & \cellcolor{gray!10}$(49.0)$\\
            \cellcolor{gray!25} & \cellcolor{gray!25} 2003 - 2006 & $50$ & $(55.8)$ &  &  & $89$ & $(56.0)$ & $51$ & $(59.0)$ & $52$ & $(58.1)$ & \cellcolor{gray!10}$192$ & \cellcolor{gray!10}$(57.4)$\\
            \cellcolor{gray!25} & \cellcolor{gray!25} 2007 - 2010 & $97$ & $(76.6)$ &  &  & $129$ & $(75.2)$ & $70$ & $(67.7)$ & $120$ & $(105.8)$ & \cellcolor{gray!10}$319$ & \cellcolor{gray!10}$(85.1)$\\
            \cellcolor{gray!25} & \cellcolor{gray!25} 2011 - 2014 & $312$ & $(76.8)$ & $1$ & $(58.0)$ & $398$ & $(78.0)$ & $222$ & $(83.3)$ & $214$ & $(89.8)$ & \cellcolor{gray!10}$834$ & \cellcolor{gray!10}$(82.4)$\\
            \cellcolor{gray!25} & \cellcolor{gray!25} 2015 - 2018 & $825$ & $(87.1)$ & $2$ & $(141.0)$ & $1178$ & $(87.7)$ & $793$ & $(88.6)$ & $714$ & $(84.5)$ & \cellcolor{gray!10}$2685$ & \cellcolor{gray!10}$(87.1)$\\
            \cellcolor{gray!25} \multirow{-7}*{\rotatebox{90}{\textbf{By Year}}} & \cellcolor{gray!25} 2019 - 2022 & $751$ & $(83.0)$ & $3$ & $(132.7)$ & $1189$ & $(84.6)$ & $1102$ & $(87.1)$ & $1037$ & $(86.5)$ & \cellcolor{gray!10}$3328$ & \cellcolor{gray!10}$(86.0)$\\
            
            \hline
            
            \cellcolor{gray!25} & \cellcolor{gray!25} DE & $250$ & $(63.7)$ &  &  & $353$ & $(63.1)$ & $260$ & $(56.9)$ & $275$ & $(52.3)$ & \cellcolor{gray!10}$888$ & \cellcolor{gray!10}$(57.9)$\\
            \cellcolor{gray!25} & \cellcolor{gray!25} DK &  &  &  &  & $1$ & $(42.0)$ &  &  &  &  & \cellcolor{gray!10}$1$ & \cellcolor{gray!10}$(42.0)$\\
            \cellcolor{gray!25} & \cellcolor{gray!25} EN & $1862$ & $(83.5)$ & $6$ & $(123.0)$ & $2729$ & $(84.8)$ & $2008$ & $(89.2)$ & $1907$ & $(90.6)$ & \cellcolor{gray!10}$6644$ & \cellcolor{gray!10}$(87.8)$\\
            \cellcolor{gray!25} & \cellcolor{gray!25} FR & $3$ & $(76.3)$ &  &  & $3$ & $(76.3)$ & $4$ & $(65.2)$ & $1$ & $(85.0)$ & \cellcolor{gray!10}$8$ & \cellcolor{gray!10}$(71.9)$\\
            \cellcolor{gray!25} & \cellcolor{gray!25} GR &  &  &  &  & $3$ & $(74.0)$ &  &  & $1$ & $(47.0)$ & \cellcolor{gray!10}$4$ & \cellcolor{gray!10}$(67.2)$\\
            \cellcolor{gray!25} \multirow{-6}*{\rotatebox{90}{\textbf{By Language}}} & \cellcolor{gray!25} IT &  &  &  &  &  &  &  &  & $1$ & $(51.0)$ & \cellcolor{gray!10}$1$ & \cellcolor{gray!10}$(51.0)$\\
            
            \hline
            \hline
            
            \multicolumn{2}{|c||}{\cellcolor{gray!25} \textbf{Total}} & \cellcolor{gray!10}$2115$ & \cellcolor{gray!10}$(81.1)$ & \cellcolor{gray!10}$6$ & \cellcolor{gray!10}$(123.0)$ & \cellcolor{gray!10}$3089$ & \cellcolor{gray!10}$(82.3)$ & \cellcolor{gray!10}$2272$ & \cellcolor{gray!10}$(85.4)$ & \cellcolor{gray!10}$2185$ & \cellcolor{gray!10}$(85.8)$ & \cellcolor{gray!10}$7546$ & \cellcolor{gray!10}$(84.2)$\\

            \hline

            \multicolumn{2}{l}{\texttt{hazard}} &
            \multicolumn{2}{c}{\tiny{salmonella}} &
            \multicolumn{2}{c}{\tiny{rhodamine b}} &
            \multicolumn{8}{c}{} \\
            
            \hline
            
            \cellcolor{gray!25} & \cellcolor{gray!25} 1994 - 1998 & $3$ & $(25.0)$ &  &  & $16$ & $(32.4)$ & $23$ & $(35.4)$ & $19$ & $(41.3)$ & \cellcolor{gray!10}$58$ & \cellcolor{gray!10}$(36.5)$\\
            \cellcolor{gray!25} & \cellcolor{gray!25} 1999 - 2002 & $6$ & $(58.7)$ &  &  & $33$ & $(48.3)$ & $46$ & $(49.1)$ & $51$ & $(49.2)$ & \cellcolor{gray!10}$130$ & \cellcolor{gray!10}$(49.0)$\\
            \cellcolor{gray!25} & \cellcolor{gray!25} 2003 - 2006 & $13$ & $(56.8)$ &  &  & $47$ & $(57.9)$ & $82$ & $(59.0)$ & $63$ & $(54.9)$ & \cellcolor{gray!10}$192$ & \cellcolor{gray!10}$(57.4)$\\
            \cellcolor{gray!25} & \cellcolor{gray!25} 2007 - 2010 & $96$ & $(105.2)$ &  &  & $138$ & $(97.9)$ & $83$ & $(72.3)$ & $98$ & $(77.9)$ & \cellcolor{gray!10}$319$ & \cellcolor{gray!10}$(85.1)$\\
            \cellcolor{gray!25} & \cellcolor{gray!25} 2011 - 2014 & $143$ & $(104.6)$ &  &  & $315$ & $(92.5)$ & $310$ & $(76.8)$ & $209$ & $(75.7)$ & \cellcolor{gray!10}$834$ & \cellcolor{gray!10}$(82.4)$\\
            \cellcolor{gray!25} & \cellcolor{gray!25} 2015 - 2018 & $288$ & $(87.1)$ & $1$ & $(95.0)$ & $1054$ & $(92.5)$ & $914$ & $(85.0)$ & $717$ & $(82.1)$ & \cellcolor{gray!10}$2685$ & \cellcolor{gray!10}$(87.1)$\\
            \cellcolor{gray!25} \multirow{-7}*{\rotatebox{90}{\textbf{By Year}}} & \cellcolor{gray!25} 2019 - 2022 & $393$ & $(84.2)$ &  &  & $1065$ & $(87.9)$ & $1006$ & $(85.6)$ & $1257$ & $(84.8)$ & \cellcolor{gray!10}$3328$ & \cellcolor{gray!10}$(86.0)$\\
            
            \hline
            
            \cellcolor{gray!25} & \cellcolor{gray!25} DE & $130$ & $(52.7)$ &  &  & $215$ & $(57.0)$ & $264$ & $(60.6)$ & $409$ & $(56.7)$ & \cellcolor{gray!10}$888$ & \cellcolor{gray!10}$(57.9)$\\
            \cellcolor{gray!25} & \cellcolor{gray!25} DK &  &  &  &  & $1$ & $(42.0)$ &  &  &  &  & \cellcolor{gray!10}$1$ & \cellcolor{gray!10}$(42.0)$\\
            \cellcolor{gray!25} & \cellcolor{gray!25} EN & $812$ & $(95.5)$ & $1$ & $(95.0)$ & $2450$ & $(92.3)$ & $2193$ & $(84.3)$ & $2001$ & $(86.1)$ & \cellcolor{gray!10}$6644$ & \cellcolor{gray!10}$(87.8)$\\
            \cellcolor{gray!25} & \cellcolor{gray!25} FR &  &  &  &  &  &  & $6$ & $(76.2)$ & $2$ & $(59.0)$ & \cellcolor{gray!10}$8$ & \cellcolor{gray!10}$(71.9)$\\
            \cellcolor{gray!25} & \cellcolor{gray!25} GR &  &  &  &  & $2$ & $(80.0)$ & $1$ & $(62.0)$ & $1$ & $(47.0)$ & \cellcolor{gray!10}$4$ & \cellcolor{gray!10}$(67.2)$\\
            \cellcolor{gray!25} \multirow{-6}*{\rotatebox{90}{\textbf{By Language}}} & \cellcolor{gray!25} IT &  &  &  &  &  &  &  &  & $1$ & $(51.0)$ & \cellcolor{gray!10}$1$ & \cellcolor{gray!10}$(51.0)$\\
            
            \hline
            \hline
            
            \multicolumn{2}{|c||}{\cellcolor{gray!25} \textbf{Total}} & \cellcolor{gray!10}$942$ & \cellcolor{gray!10}$(89.6)$ & \cellcolor{gray!10}$1$ & \cellcolor{gray!10}$(95.0)$ & \cellcolor{gray!10}$2668$ & \cellcolor{gray!10}$(89.4)$ & \cellcolor{gray!10}$2464$ & \cellcolor{gray!10}$(81.8)$ & \cellcolor{gray!10}$2414$ & \cellcolor{gray!10}$(81.1)$ & \cellcolor{gray!10}$7546$ & \cellcolor{gray!10}$(84.2)$\\

            \hline

            \multicolumn{2}{l}{\texttt{product}} &
            \multicolumn{2}{c}{\tiny{ice cream}} &
            \multicolumn{2}{c}{\tiny{canned meat stew}} &
            \multicolumn{8}{c}{} \\
            
            \hline
            
            \cellcolor{gray!25} & \cellcolor{gray!25} 1994 - 1998 & $3$ & $(48.3)$ &  &  & $34$ & $(37.0)$ & $18$ & $(36.9)$ & $6$ & $(32.3)$ & \cellcolor{gray!10}$58$ & \cellcolor{gray!10}$(36.5)$\\
            \cellcolor{gray!25} & \cellcolor{gray!25} 1999 - 2002 & $8$ & $(60.0)$ &  &  & $62$ & $(49.5)$ & $36$ & $(49.3)$ & $32$ & $(47.4)$ & \cellcolor{gray!10}$130$ & \cellcolor{gray!10}$(49.0)$\\
            \cellcolor{gray!25} & \cellcolor{gray!25} 2003 - 2006 & $6$ & $(61.7)$ &  &  & $63$ & $(59.3)$ & $71$ & $(57.1)$ & $58$ & $(55.7)$ & \cellcolor{gray!10}$192$ & \cellcolor{gray!10}$(57.4)$\\
            \cellcolor{gray!25} & \cellcolor{gray!25} 2007 - 2010 & $69$ & $(120.7)$ &  &  & $163$ & $(97.4)$ & $81$ & $(70.1)$ & $75$ & $(74.6)$ & \cellcolor{gray!10}$319$ & \cellcolor{gray!10}$(85.1)$\\
            \cellcolor{gray!25} & \cellcolor{gray!25} 2011 - 2014 & $45$ & $(120.8)$ &  &  & $331$ & $(92.2)$ & $261$ & $(75.0)$ & $242$ & $(77.1)$ & \cellcolor{gray!10}$834$ & \cellcolor{gray!10}$(82.4)$\\
            \cellcolor{gray!25} & \cellcolor{gray!25} 2015 - 2018 & $64$ & $(100.0)$ &  &  & $930$ & $(88.3)$ & $861$ & $(87.4)$ & $894$ & $(85.7)$ & \cellcolor{gray!10}$2685$ & \cellcolor{gray!10}$(87.1)$\\
            \cellcolor{gray!25} \multirow{-7}*{\rotatebox{90}{\textbf{By Year}}} & \cellcolor{gray!25} 2019 - 2022 & $57$ & $(97.4)$ & $1$ & $(90.0)$ & $925$ & $(90.6)$ & $1227$ & $(84.4)$ & $1176$ & $(84.1)$ & \cellcolor{gray!10}$3328$ & \cellcolor{gray!10}$(86.0)$\\
            
            \hline
            
            \cellcolor{gray!25} & \cellcolor{gray!25} DE & $6$ & $(93.5)$ &  &  & $196$ & $(63.9)$ & $320$ & $(55.4)$ & $372$ & $(57.0)$ & \cellcolor{gray!10}$888$ & \cellcolor{gray!10}$(57.9)$\\
            \cellcolor{gray!25} & \cellcolor{gray!25} DK &  &  &  &  &  &  & $1$ & $(42.0)$ &  &  & \cellcolor{gray!10}$1$ & \cellcolor{gray!10}$(42.0)$\\
            \cellcolor{gray!25} & \cellcolor{gray!25} EN & $246$ & $(106.3)$ & $1$ & $(90.0)$ & $2309$ & $(89.9)$ & $2229$ & $(86.4)$ & $2106$ & $(87.0)$ & \cellcolor{gray!10}$6644$ & \cellcolor{gray!10}$(87.8)$\\
            \cellcolor{gray!25} & \cellcolor{gray!25} FR &  &  &  &  & $2$ & $(87.0)$ & $1$ & $(73.0)$ & $5$ & $(65.6)$ & \cellcolor{gray!10}$8$ & \cellcolor{gray!10}$(71.9)$\\
            \cellcolor{gray!25} & \cellcolor{gray!25} GR &  &  &  &  & $1$ & $(62.0)$ & $3$ & $(69.0)$ &  &  & \cellcolor{gray!10}$4$ & \cellcolor{gray!10}$(67.2)$\\
            \cellcolor{gray!25} \multirow{-6}*{\rotatebox{90}{\textbf{By Language}}} & \cellcolor{gray!25} IT &  &  &  &  &  &  & $1$ & $(51.0)$ &  &  & \cellcolor{gray!10}$1$ & \cellcolor{gray!10}$(51.0)$\\
            
            \hline
            \hline
            
            \multicolumn{2}{|c||}{\cellcolor{gray!25} \textbf{Total}} & \cellcolor{gray!10}$252$ & \cellcolor{gray!10}$(106.0)$ & \cellcolor{gray!10}$1$ & \cellcolor{gray!10}$(90.0)$ & \cellcolor{gray!10}$2508$ & \cellcolor{gray!10}$(87.9)$ & \cellcolor{gray!10}$2555$ & \cellcolor{gray!10}$(82.4)$ & \cellcolor{gray!10}$2483$ & \cellcolor{gray!10}$(82.4)$ & \cellcolor{gray!10}$7546$ & \cellcolor{gray!10}$(84.2)$\\

            \hline
        \end{tabular}
    }
    \caption{Statistical overview of the data by prediction task. Each cell contains the number of samples in this category as well as the average text length of these samples in characters (in brackets). Empty cells indicate no samples in this category.}
    \label{tab:data_statistics}
\end{table*}

\begin{figure*}[hb]
    \centering
    \begin{tabular}{|r|rl|rl|}
    \hline
    
    \multicolumn{5}{|p{.95\linewidth}|}{\cellcolor{gray!20}\footnotesize "\colorbox{blue!40}{Randsland brand Super \colorbox{orange!40}{Salad}} Kit recalled due to \colorbox{orange!40}{Listeria} monocytogenes"} \\
    \hline
    \multirow{2}{*}{Labels}
    & \footnotesize\texttt{\color{orange}hazard}:           & \footnotesize listeria monocytogenes
    & \footnotesize\texttt{\color{orange}hazard-category}:  & \footnotesize biological \\
    & \footnotesize\texttt{\color{blue}product}:            & \footnotesize salads
    & \footnotesize\texttt{\color{blue}product-category}:   & \footnotesize fruits and vegetables \\

    \hline
    \hline
    
    \multicolumn{5}{|p{.95\linewidth}|}{\cellcolor{gray!20}\footnotesize "Create Common Good Recalls Jambalaya \colorbox{blue!40}{\colorbox{orange!40}{Products}} Due To Misbranding and Undeclared \colorbox{orange!40}{Allergens}"} \\
    \hline
    \multirow{2}{*}{Labels}
    & \footnotesize\texttt{\color{orange}hazard}:           & \footnotesize milk and products thereof
    & \footnotesize\texttt{\color{orange}hazard-category}:  & \footnotesize allergens \\
    & \footnotesize\texttt{\color{blue}product}:            & \footnotesize meat preparations
    & \footnotesize\texttt{\color{blue}product-category}:   & \footnotesize meat, egg and dairy products \\

    \hline
    \hline
    
    \multicolumn{5}{|p{.95\linewidth}|}{\cellcolor{gray!20}\footnotesize "Revised Guan’s \colorbox{blue!40}{\colorbox{orange!40}{Mushroom}} Co Recalls \colorbox{blue!40}{\colorbox{orange!40}{Enoki}} Because of Possible \colorbox{orange!40}{Health Risk}"} \\
    \hline
    \multirow{2}{*}{Labels}
    & \footnotesize\texttt{\color{orange}hazard}:           & \footnotesize listeria monocytogenes
    & \footnotesize\texttt{\color{orange}hazard-category}:  & \footnotesize biological \\
    & \footnotesize\texttt{\color{blue}product}:            & \footnotesize mushrooms
    & \footnotesize\texttt{\color{blue}product-category}:   & \footnotesize fruits and vegetables \\

    \hline
    \hline
    
    \multicolumn{5}{|p{.95\linewidth}|}{\cellcolor{gray!20}\footnotesize "Nestlé Prepared Foods Recalls Lean Cuisine Baked \colorbox{blue!40}{Chicken Meal Products} Due to Possible \colorbox{blue!40}{\colorbox{orange!40}{Foreign}} Matter Contamination"} \\
    \hline
    \multirow{2}{*}{Labels}
    & \footnotesize\texttt{\color{orange}hazard}:           & \footnotesize plastic fragment
    & \footnotesize\texttt{\color{orange}hazard-category}:  & \footnotesize foreign bodies \\
    & \footnotesize\texttt{\color{blue}product}:            & \footnotesize cooked chicken
    & \footnotesize\texttt{\color{blue}product-category}:   & \footnotesize prepared dishes and snacks \\

    \hline
    \hline
    
    \multicolumn{5}{|p{.95\linewidth}|}{\cellcolor{gray!20}\footnotesize "Undeclared allergen (\colorbox{blue!40}{sulphur} dioxide and sorbic \colorbox{orange!40}{acid} ) found in prepackaged \colorbox{blue!40}{date} sample"} \\
    \hline
    \multirow{2}{*}{Labels}
    & \footnotesize\texttt{\color{orange}hazard}:           & \footnotesize too high content of e 200 - sorbic acid
    & \footnotesize\texttt{\color{orange}hazard-category}:  & \footnotesize food additives and flavourings \\
    & \footnotesize\texttt{\color{blue}product}:            & \footnotesize dates
    & \footnotesize\texttt{\color{blue}product-category}:   & \footnotesize fruits and vegetables \\

    \hline
    \end{tabular}
    \caption{Some labeled sample texts. Colored spans signify the spans in \texttt{\color{orange}hazard-title} and \texttt{\color{blue}product-title}}
    \label{fig:example_texts}
\end{figure*}

\FloatBarrier
\begin{table*}[ht]
\section{Sample Prompts:}\label{sec:appendix_prompts}

\hspace{5pt}

\centering
\resizebox*{\textwidth}{!}{
    \begin{tabular}{|ll|}
        \hline

        \multicolumn{2}{|l|}{\cellcolor{gray!25}\textbf{GPT-ALL:}} \\
        \hline

        & \textbf{We are looking for food hazards in texts. Please predict the correct class for the following sample:} \\
        & \color{RoyalBlue}{"CSM Bakery Solutions Voluntary Recalls 8" => allergens} \\
        & \color{RoyalBlue}{"CSM Bakery Solutions' Voluntary Recall of Cinnabon Stix® Due to Undeclared Peanut Allergen" => allergens} \\
        & \color{Orange}{"H-E-B Issues Voluntary Recall On Bakery Products" => biological} \\
        & \color{Orange}{"2009 - christie cookie recalls certain lots of peanut butter cookies due to expanded recall by peanut corporation of america" => biological} \\
        & \color{Green}{"Gourmet Culinary Solutions Recalls Turkey Sausage Products Due To Possible Foreign Matter Contamination" => foreign bodies} \\
        & \color{Magenta}{"Golden Natural Products Inc. Issues Allergy Alert on Undeclared Sulfites in Two Dried Apricot Products" => fraud} \\
        & \color{Magenta}{"Texas Best Protein DBA Farm to Market Foods Issues Allergy Alert on Undeclared Peanut in Green Bean Casserole" => fraud} \\
        & \color{Green}{"SPAR recalls its SPAR Chicken Tikka Chunks because it may contain small pieces of glass" => foreign bodies} \\
        & \color{Red}{"Lactalis Nestlé Chilled Dairy UK recalls Nescafé Shakissimo Espresso Latte because of contamination with cleaning solution" => chemical} \\
        & \color{Red}{"oriental packing co. inc. issues alert on lead in curry powder" => chemical} \\
        & \color{Cyan}{"Garden Fortune Cookies" => migration} \\
        & \color{Gray}{"Green Field Farms Dairy Issues Voluntary Recall of Whole Chocolate Milk" => other hazard} \\
        & \color{Purple}{"Bakery Product - Labelling - Undeclared food additives (Preservative (211) and Colour (129) - 29 November 2010" => food additives and flavourings} \\
        & \color{Gray}{"FSIS Issues Public Health Alert for Ready-To-Eat Ham Product Due to Possible Processing Deviation" => other hazard} \\
        & \color{Turquoise}{"Pauls—Breaka chocolate UHT milk" => organoleptic aspects} \\
        & \color{Turquoise}{"Nestle—Billabong chocolate ice cream 8 pack" => organoleptic aspects} \\
        & \color{Purple}{"Undeclared allergen (sulphur dioxide and sorbic acid) found in prepackaged date sample" => food additives and flavourings} \\
        & \color{PineGreen}{"Heinz—Baked Beans in Tomato Sauce" => packaging defect} \\
        & \color{PineGreen}{"John West—Scottish sardines In vegetable oil" => packaging defect} \\
        & \color{Cyan}{"Notos Schafskäse / Bulgarischer Schafskäse in Salzlake gereift 200g" => migration} \\
        & \textbf{"CSM Bakery Solutions Issues Allergy Alert on Undeclared Peanut in Chick-Fil-A Chocolate Chunk Cookies" => } \\

        \hline
        \hline

        \multicolumn{2}{|l|}{\cellcolor{gray!25}\textbf{GPT-SIM-$k$:}} \\
        \hline

        &\textbf {We are looking for food hazards in texts. Please predict the correct class for the following sample:} \\

        \color{gray}{$k=\enspace 1 \rightarrow$}
        & \color{RoyalBlue}{"CSM Bakery Solutions Voluntary Recalls 8" => allergens} \\

        \color{gray}{$k=\enspace 2 \rightarrow$}
        & \color{RoyalBlue}{"CSM Bakery Solutions' Voluntary Recall of Cinnabon Stix® Due to Undeclared Peanut Allergen" => allergens} \\

        \color{gray}{$k=\enspace 3 \rightarrow$}
        & \color{Orange}{"H-E-B Issues Voluntary Recall On Bakery Products" => biological} \\

        \color{gray}{$k=\enspace 4 \rightarrow$}
        & \color{Orange}{"2009 - christie cookie recalls certain lots of peanut butter cookies due to expanded recall by peanut corporation of america" => biological} \\

        \color{gray}{$k=\enspace 5 \rightarrow$}
        & \color{Green}{"Gourmet Culinary Solutions Recalls Turkey Sausage Products Due To Possible Foreign Matter Contamination" => foreign bodies} \\

        \color{gray}{$k=\enspace 6 \rightarrow$}
        & \color{Magenta}{"Golden Natural Products Inc. Issues Allergy Alert on Undeclared Sulfites in Two Dried Apricot Products" => fraud} \\

        \color{gray}{$k=\enspace 7 \rightarrow$}
        & \color{Magenta}{"Texas Best Protein DBA Farm to Market Foods Issues Allergy Alert on Undeclared Peanut in Green Bean Casserole" => fraud} \\

        \color{gray}{$k=\enspace 8 \rightarrow$}
        & \color{Green}{"SPAR recalls its SPAR Chicken Tikka Chunks because it may contain small pieces of glass" => foreign bodies} \\

        \color{gray}{$k=\enspace 9 \rightarrow$}
        & \color{Red}{"Lactalis Nestlé Chilled Dairy UK recalls Nescafé Shakissimo Espresso Latte because of contamination with cleaning solution" => chemical} \\

        \color{gray}{$k=10 \rightarrow$}
        & \color{Red}{"oriental packing co. inc. issues alert on lead in curry powder" => chemical} \\

        & \textbf{"CSM Bakery Solutions Issues Allergy Alert on Undeclared Peanut in Chick-Fil-A Chocolate Chunk Cookies" => } \\

        \hline
        \hline

        \multicolumn{2}{|l|}{\cellcolor{gray!25}\textbf{GPT-MAX-$k$:}} \\
        \hline

        & \textbf{We are looking for food hazards in texts. Please predict the correct class for the following sample:} \\
        
        & \color{RoyalBlue}{"CSM Bakery Solutions Voluntary Recalls 8" => allergens} \\
        \color{gray}{$k=1 \rightarrow$}
        & \color{RoyalBlue}{"CSM Bakery Solutions' Voluntary Recall of Cinnabon Stix® Due to Undeclared Peanut Allergen" => allergens} \\
        
        & \color{Magenta}{"Golden Natural Products Inc. Issues Allergy Alert on Undeclared Sulfites in Two Dried Apricot Products" => fraud} \\
        \color{gray}{$k=2 \rightarrow$}
        & \color{Magenta}{"Texas Best Protein DBA Farm to Market Foods Issues Allergy Alert on Undeclared Peanut in Green Bean Casserole" => fraud} \\
        
        & \color{PineGreen}{"Heinz—Baked Beans in Tomato Sauce" => packaging defect} \\
        \color{gray}{$k=3 \rightarrow$}
        & \color{PineGreen}{"John West—Scottish sardines In vegetable oil" => packaging defect} \\
        
        & \color{Turquoise}{"Pauls—Breaka chocolate UHT milk" => organoleptic aspects} \\
        \color{gray}{$k=4 \rightarrow$}
        & \color{Turquoise}{"Nestle—Billabong chocolate ice cream 8 pack" => organoleptic aspects} \\
        
        & \color{Gray}{"Green Field Farms Dairy Issues Voluntary Recall of Whole Chocolate Milk" => other hazard} \\
        \color{gray}{$k=5 \rightarrow$}
        & \color{Gray}{"FSIS Issues Public Health Alert for Ready-To-Eat Ham Product Due to Possible Processing Deviation" => other hazard} \\
        
        & \textbf{"CSM Bakery Solutions Issues Allergy Alert on Undeclared Peanut in Chick-Fil-A Chocolate Chunk Cookies" => } \\

        \hline
        \hline

        \multicolumn{2}{|l|}{\cellcolor{gray!25}\textbf{GPT-CICLe:}} \\
        \hline

        & \textbf{We are looking for food hazards in texts. Please predict the correct class for the following sample:} \\
        & \color{RoyalBlue}{"CSM Bakery Solutions Voluntary Recalls 8" => allergens} \\
        & \color{RoyalBlue}{"CSM Bakery Solutions' Voluntary Recall of Cinnabon Stix® Due to Undeclared Peanut Allergen" => allergens} \\
        & \textbf{"CSM Bakery Solutions Issues Allergy Alert on Undeclared Peanut in Chick-Fil-A Chocolate Chunk Cookies" => } \\

        \hline
    \end{tabular}
}
\caption{Sample prompts taken from the \texttt{hazard-category} classification task. Example classes are color coded for better visibility. The order of the examples in GPT-ALL and GPT-SIM-$k$ is identical, while in GPT-CICLe and GPT-MAX-$k$ they are sorted by model certainty (most probable one first).}
\label{tab:prompts}
\end{table*}

\FloatBarrier

\section{Model Specifics}\label{sec:appendix_models}

\subsection{Embedings}
As discussed in §\ref{sec:method}, we use two different types of embeddings for our classical models: \textit{bag-of-words}~(BOW) and \textit{tf-idf}~(TF-IDF).
For the BOW representation,\footnote{We use or own implementation for the representations.} we create a vector:
\[X_i=[x_{i,1}, x_{i,2}, \dots, x_{i,V}],~\forall x_{i,v} \in [0,1]\]
for each sample~$i$, where $V$ is the vocabulary size, and $x_{i,v}$ is the normalized count of occurrences of token~$v$ in sample~$i$. For the TF-IDF-embedding, we calcluate
\[x_{i,v} = \mathrm{tf}_{i,v}\cdot\mathrm{idf}_v,\]
where $\mathrm{tf}_{i,v}$ is the count of the samples' $v$-th token in the $i$-th training sample, and $\mathrm{idf}_v=\ln{\frac{N}{n_v}}$ with $n_v$ being the number of documents that contain token~$v$.
In order to make the above embeddings independent of the sample length we normalize $X_i$ with the L2-norm in both cases.
Text pre-processing comprises the application of a \texttt{TreebankWordTokenizer}, followed by \texttt{PorterStemmer} from the \texttt{NLTK}~\cite{nltk} Python package.

\subsection{Traditional Classifiers}

\begin{table}[h]
    \centering
    \resizebox*{\linewidth}{!}{
    
    \begin{tabular}{lllll}
    \hline
    \textbf{Model} &             \textbf{Task} &                                \textbf{Parameters} \\
    \hline
    
    \multirow{4}{*}{BOW-KNN}
    & \texttt{hazard-category}  & [$k=2$]~($3\times$) + [$k=4$]~($2\times$) \\
    & \texttt{product-category} & [$k=2$]~($4\times$) + [$k=4$]~($1\times$) \\
    & \texttt{hazard}           & [$k=2$]~($3\times$) + [$k=4$]~($2\times$) \\
    & \texttt{product}          & [$k=2$]~($3\times$) + [$k=8$]~($2\times$) \\
    
    \hline
    
    \multirow{7}{*}{TFIDF-KNN}
    & \texttt{hazard-category}                   & [$k=4$]~($1\times$) + [$k=8$]~($4\times$) \\
    & \multirow{2}{*}{\texttt{product-category}} & [$k=2$]~($2\times$) + [$k=4$]~($2\times$) +\\
    &                                            & [$k=8$]~($1\times$) \\
    & \multirow{2}{*}{\texttt{hazard}}           & [$k=2$]~($1\times$) + [$k=4$]~($1\times$) +\\
    &                                            & [$k=8$]~($3\times$) \\
    & \multirow{2}{*}{\texttt{product}}          & [$k=2$]~($1\times$) + [$k=4$]~($1\times$) +\\
    &                                            & [$k=8$]~($3\times$) \\
    
    \hline
    
    \multirow{4}{*}{BOW-LR}
    & \texttt{hazard-category}  & [L1, $C=2.0$]~($5\times$) \\
    & \texttt{product-category} & [L1, $C=2.0$]~($5\times$) \\
    & \texttt{hazard}           & [L1, $C=2.0$]~($5\times$) \\
    & \texttt{product}          & [L1, $C=2.0$]~($5\times$) \\
    
    \hline
    
    \multirow{4}{*}{TFIDF-LR}
    & \texttt{hazard-category}  & [L1, $C=2.0$]~($5\times$) \\
    & \texttt{product-category} & [L1, $C=2.0$]~($5\times$) \\
    & \texttt{hazard}           & [L1, $C=2.0$]~($5\times$) \\
    & \texttt{product}          & [L1, $C=2.0$]~($5\times$) \\
    
    \hline
    
    \multirow{4}{*}{BOW-SVM}
    & \texttt{hazard-category}  & [$C=2.0$]~($5\times$) \\
    & \texttt{product-category} & [$C=2.0$]~($5\times$) \\
    & \texttt{hazard}           & [$C=2.0$]~($5\times$) \\
    & \texttt{product}          & [$C=2.0$]~($5\times$) \\
    
    \hline
    
    \multirow{4}{*}{TFIDF-SVM}
    & \texttt{hazard-category}  & [$C=1.0$]~($1\times$) + [$C=2.0$]~($4\times$) \\
    & \texttt{product-category} & [$C=2.0$]~($5\times$) \\
    & \texttt{hazard}           & [$C=2.0$]~($5\times$) \\
    & \texttt{product}          & [$C=2.0$]~($5\times$) \\
    
    \hline
    \end{tabular}
    }
    \caption{Parameters of Traditional Classifiers}
    \label{tab:params_traditional}
\end{table}

As already discussed in §\ref{sec:method}, we optimze the models trained on each of our $5$ CV splits on a calibration set created using $10\%$ holdout on the respective training data. We optimize...
\begin{tabular}{r p{6cm}}
     ...&$k\in\{2, 4, 8\}$ neighbours for KNN \\
     ...&$C \in \{0.5, 1.0, 2.0\}$ for both L1 and L2 for LR \\
     ...&$C \in \{0.5, 1.0, 2.0\}$ for SVM \\
\end{tabular}
The best values found for each of the splits are reported in Table~\ref{tab:params_traditional}. Apart from KNN, we find stable sets of parameters for the best performance during grid search.

\subsection{Encoder-only-Transformers}

\begin{figure}[hb]
    \centering
    \includegraphics[width=\linewidth]{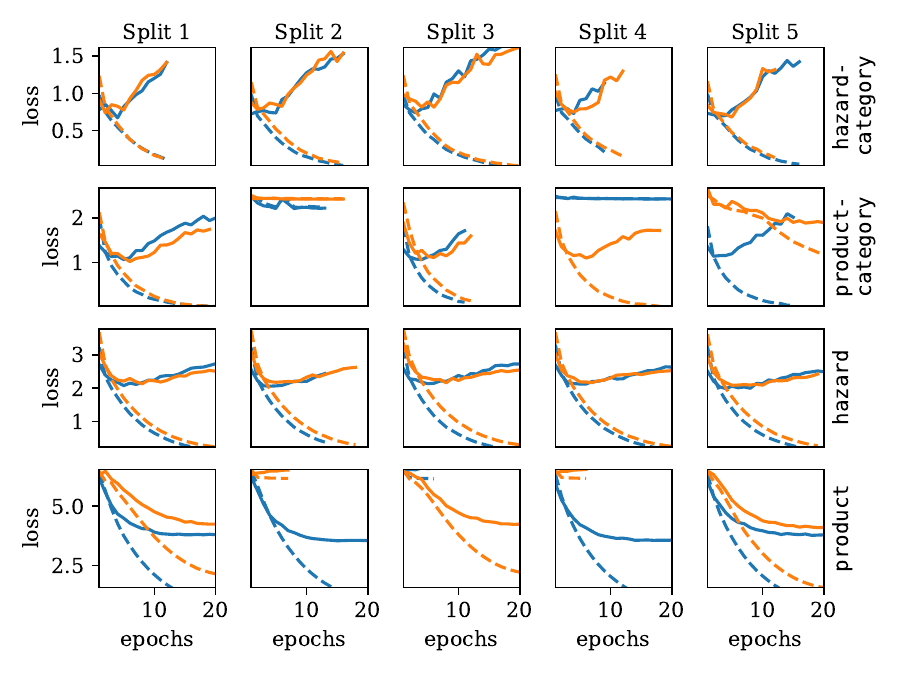}
    \includegraphics[width=.4\linewidth, trim={.9cm 1cm .5cm .5cm}, clip]{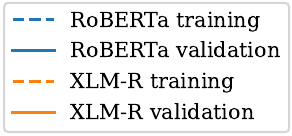}
    \caption{Learning curves of fine-tuning of RoBERTa and XLM-R averaged over 5 CV-splits.}
    \label{fig:params_transformers}
\end{figure}

\noindent For fine-tuning RoBERTa and XLM-R we use a learning rate of $5 \cdot 10^{-5}$ as well as a batch size of $16$. Instead of grid searching for the best parameter we employ early stopping at the minimum validation loss with a patience of $5$~epochs. As the average learning curves in Figure~\ref{fig:params_transformers} show, \texttt{product-category} and \texttt{product} are harder to learn, as for some of the splits the curves show no progression. For the remaining models we reach the mininmum validation loss at an epoch of 5 on average. 

\subsection{CICLe}
As we are aiming for a low set size, we select the $\alpha$ based on Figure~\ref{fig:conformal_performance}. For \texttt{hazard-category} and \texttt{product-category}, we choose $\alpha=0.05$ (meaning $p=0.95$), as even for this high accuracy we get set sizes smaller than $10$. For the remaining classification tasks we are limited by set length and choose $\alpha=0.20$ for \texttt{hazard} and $\alpha=0.40$ for \texttt{product}. This means, that for \texttt{product} we have to accept a chance of at most $40\%$ that the prediction set does not contain the ground truth in order to keep the set size low.
This results in on average $2.6$ classes per prompt in \texttt{hazard-category}, $5.3$ classes per prompt in \texttt{product-category}, $4.7$ classes per prompt in \texttt{hazard}, and $6.7$ classes per prompt in \texttt{product}.

\FloatBarrier

\section{Extended Results}\label{sec:appendix_results}

In this appendix, we present additional results not included in the main paper. Tables~\ref{tab:results_extended} and \ref{tab:results_prompting_extended} show additional metrics (\textit{micro}-$F_1$, \textit{macro}-precision, and \textit{macro}-recall), as well as the bag-of-word~(BOW) based classifiers.
Table~\ref{tab:langauge_based_assessment} compares \textit{macro}-$F_1$ of RoBERTa and XLM-R on the two most represented languages in our data: English and German. We refer to §\ref{sec:analysis} for a discussion of the aforementioned tables.

\subsection{Theoretical boundaries on prompting performance}
In Table~\ref{tab:results_prompting_limits} we present upper and lower bounds on the performance of our prompting strategies. In order to estimate the upper bounds~(*-MAX), we assume that the LLM predicts the true class every time it is included in the prompt, and otherwise, the class of a random sample in the prompt. As is to be expected, the highest upper bound is always taken by the *-ALL prompts, as they include all classes in the training set. Note that for \texttt{hazard} and \texttt{product} the maximum scores are $< 1$, indicating that not all classes are present in the training data. This is a consequence of the low class support for some of the classes, which can not be divided into the five CV-splits.
On our data, the reduction mostly affects the low-support classes, as they make up the bulk of the data but have a low number of tetxs per class. This means there is a higher possibility of completely removing them from the prompt compared to the other segments. Maximum performance on high support classes remains almost untouched by the reduction as they are very unlikely to be completely removed from the prompts.

For the lower bounds~(MIN-*), we simply use the class of a random few-shot-sample in each prompt as the prediction. Here we assume that an LLM never predicts worse than random. The best minimum performance is delivered by CICLe most of the time, indicating that it produces short prompts with a high probability of containing the true class.

\subsection{Analysis of prompting failures}
We also provide a qualitative assessment of why prompting fails to produce a valid class label in some cases. Based on our experiments, we identified three reasons for failures:
\begin{enumerate}

    \vspace{-5pt}
    \item Insufficient information in the sample for determining a class. In such cases, we found that the output of GPT explicitly states the problem.
    For example, consider the following texts and the corresponding LLM outputs for prompting on the \texttt{hazard} task: 
    \begin{tabular}{rp{5cm}}
        \hline
        \footnotesize\textbf{text:}      & \footnotesize "Li Li Handmade Chicken and Pork Dumplings" \\
        \footnotesize\textbf{output:}    & \footnotesize "none (no food hazards mentioned)" \\
        \hline
        \footnotesize\textbf{text:}      & \footnotesize "The Double Cola Company Recalls Select Cases of Its Cherry Ski Product"  \\
        \footnotesize\textbf{output:}    & \footnotesize "none"  \\
        \hline
        \footnotesize\textbf{text:}      & \footnotesize "The Original Smoke and Spice Co Sea Salt Gourmet Seasoning" \\
        \footnotesize\textbf{output:}    & \footnotesize "none (no food hazards mentioned)" \\
        \hline
    \end{tabular}
    None of the above examples contain a food hazard. GPT therefore states the lack of evidence which does not lead to a valid class label. We argue that in a real-world scenario, this is a strength rather than a weakness, as it can help avoid misclassification.

    \vspace{-5pt}
    \item The text contains more than one possible class. In such cases, we found that the LLM produces a mix of labels:
    \begin{tabular}{rp{5cm}}
        \hline
        \footnotesize\textbf{text:}      & \footnotesize "Bravo Packing, Inc. Recalls All Performance Dog and Ground Beef Raw Pet Food Because of Possible Salmonella and Listeria Monocytogenes Health Risk to Humans and Animals" \\
        \footnotesize\textbf{class:}     & \footnotesize "listeria monocytogenes" \\
        \footnotesize\textbf{output:}    & \footnotesize "salmonella and listeria monocytogenes" \\
        \hline
        \footnotesize\textbf{text:}      & \footnotesize "Exceptional Health Products Issues Allergy Alert On Undeclared Soy And Milk Allergens In Angel Wings 99- Daily Multi 120 Capsules" \\
        \footnotesize\textbf{class:}     & \footnotesize "milk and products thereof" \\
        \footnotesize\textbf{output:}    & \footnotesize "soybeans and products thereof, milk and products thereof"\\
        \hline
        \footnotesize\textbf{text:}      & \footnotesize "Great Value brand Spaghetti Marinara recalled due to undeclared milk and sulphites" \\
        \footnotesize\textbf{class:}     & \footnotesize "milk and products thereof" \\
        \footnotesize\textbf{output:}    & \footnotesize "milk and products thereof, sulphur dioxide and sulphites" \\
        \hline
    \end{tabular}
    The above examples are again for prompting on the \texttt{hazard} task. The texts contain multiple food hazards, which are combined by the LLM. Again, this can be seen as a strength of prompting rather than a weakness.

    \item Too few few-shot-examples per class in the prompt. In Table~\ref{tab:results_prompting} we see a high failure rate for the GPT-SIM-$k$ prompts in \texttt{product}. A closer look at GPT's failed responses to these prompts shows the following:
    \begin{tabular}{rp{5cm}}
        \hline
        \footnotesize\textbf{text:}      & \footnotesize "New American Food Products LLC Issues Allergy Alert on Undeclared Milk in Packaged Goods Received From Our Supplier GKI Foods." \\
        \footnotesize\textbf{class:}     & \footnotesize "dark chocolate coated almonds" \\
        \footnotesize\textbf{output:}    & \footnotesize "dairy products" \\
        \hline
        \footnotesize\textbf{text:}      & \footnotesize "Keats London Vegan Irish Cream Truffles 140g" \\
        \footnotesize\textbf{class:}     & \footnotesize "chocolate truffles" \\
        \footnotesize\textbf{output:}    & \footnotesize "vegan truffles" \\
        \hline
        \footnotesize\textbf{text:}      & \footnotesize "New York Firm Recalls Boneless Veal Products Due To Possible E. Coli Contamination" \\
        \footnotesize\textbf{class:}     & \footnotesize "veal" \\
        \footnotesize\textbf{output:}    & \footnotesize "veal products" \\
        \hline
    \end{tabular}
    In these cases, the LLM is not able to reproduce the exact label although it often finds the right product. We argue that because of the low number of samples per class, the LLM is not able to capture the context.

\end{enumerate}
The above samples show, that prompting failures are not necessarily a sign of bad performance (see cases 1. and 2.). In cases where the model is unable to capture the context (e.g. case 3.), better prompt design can help reduce the failure rate.

\subsection{Confusion Matrices:}\label{sec:appendix_cm}

\begin{figure}[ht]
    \centering

    \begin{tikzpicture}

        \draw (3.1,1) node[anchor=north]{\textsc{Predicted Class}};
        \draw (-1,-6.2) node[anchor=west]{\rotatebox{90}{\textsc{True Class}}};

        \draw (0.5,.5) node[anchor=north]{\footnotesize High};
        \draw (1.5,.5) node[anchor=north]{\footnotesize Medium};
        \draw (2.5,.5) node[anchor=north]{\footnotesize Low};

        \draw (3.7,.5) node[anchor=north]{\footnotesize High};
        \draw (4.7,.5) node[anchor=north]{\footnotesize Medium};
        \draw (5.7,.5) node[anchor=north]{\footnotesize Low};

        \draw (-.5,-0.5) node[anchor=west]{\rotatebox{90}{\footnotesize High}};
        \draw (-.5,-1.5) node[anchor=west]{\rotatebox{90}{\footnotesize Medium}};
        \draw (-.5,-2.5) node[anchor=west]{\rotatebox{90}{\footnotesize Low}};

        \draw (1.5,-1.5) node[anchor=center]{\includegraphics[height=3cm, trim={.95cm .8cm .75cm .9cm}, clip]{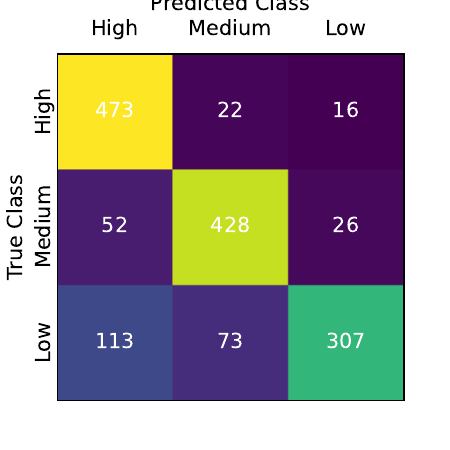}};
        \draw (4.7,-1.5) node[anchor=center]{\includegraphics[height=3cm, trim={.95cm .8cm .75cm .9cm}, clip]{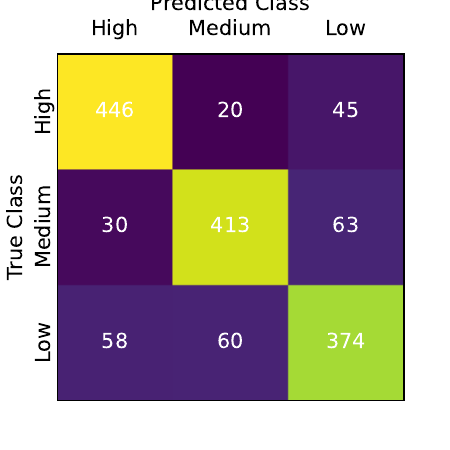}};

        \draw (6.3,-1.5) node[anchor=west]{\rotatebox{90}{\texttt{hazard-category}}};

        \draw (-.5,-3.7) node[anchor=west]{\rotatebox{90}{\footnotesize High}};
        \draw (-.5,-4.7) node[anchor=west]{\rotatebox{90}{\footnotesize Medium}};
        \draw (-.5,-5.7) node[anchor=west]{\rotatebox{90}{\footnotesize Low}};

        \draw (1.5,-4.7) node[anchor=center]{\includegraphics[height=3cm, trim={.95cm .8cm .75cm .9cm}, clip]{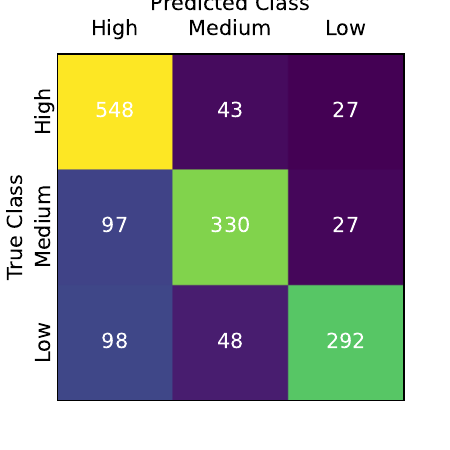}};
        \draw (4.7,-4.7) node[anchor=center]{\includegraphics[height=3cm, trim={.95cm .8cm .75cm .9cm}, clip]{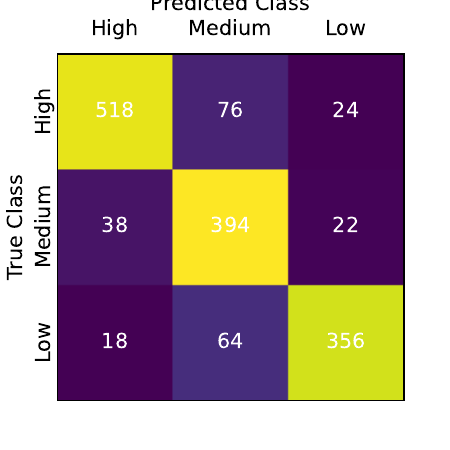}};

        \draw (6.3,-4.7) node[anchor=west]{\rotatebox{90}{\texttt{product-category}}};

        \draw (-.5,-6.9) node[anchor=west]{\rotatebox{90}{\footnotesize High}};
        \draw (-.5,-7.9) node[anchor=west]{\rotatebox{90}{\footnotesize Medium}};
        \draw (-.5,-8.9) node[anchor=west]{\rotatebox{90}{\footnotesize Low}};

        \draw (1.5,-7.9) node[anchor=center]{\includegraphics[height=3cm, trim={.95cm .8cm .75cm .9cm}, clip]{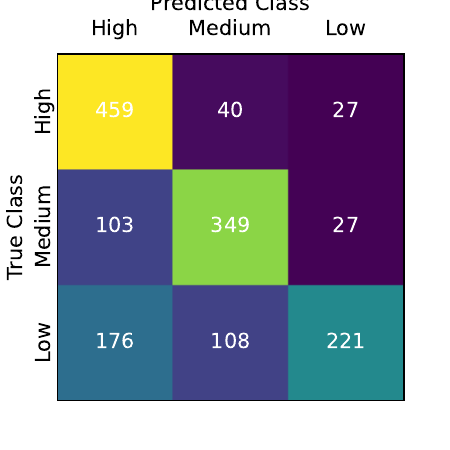}};
        \draw (4.7,-7.9) node[anchor=center]{\includegraphics[height=3cm, trim={.95cm .8cm .75cm .9cm}, clip]{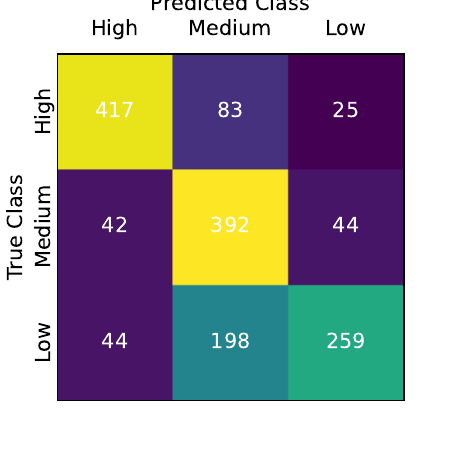}};

        \draw (6.3,-7.9) node[anchor=west]{\rotatebox{90}{\texttt{hazard}}};

        \draw (-.5,-10.1) node[anchor=west]{\rotatebox{90}{\footnotesize High}};
        \draw (-.5,-11.1) node[anchor=west]{\rotatebox{90}{\footnotesize Medium}};
        \draw (-.5,-12.1) node[anchor=west]{\rotatebox{90}{\footnotesize Low}};

        \draw (1.5,-11.1) node[anchor=center]{\includegraphics[height=3cm, trim={.95cm .8cm .75cm .9cm}, clip]{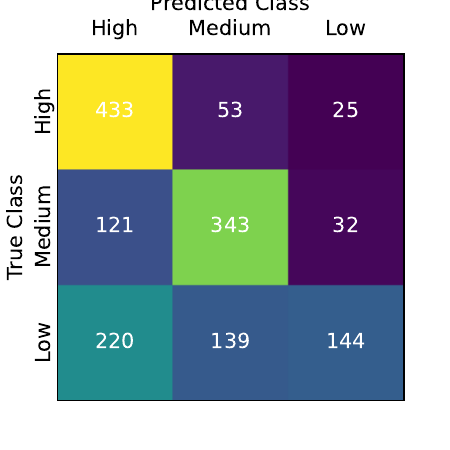}};
        \draw (4.7,-11.1) node[anchor=center]{\includegraphics[height=3cm, trim={.95cm .8cm .75cm .9cm}, clip]{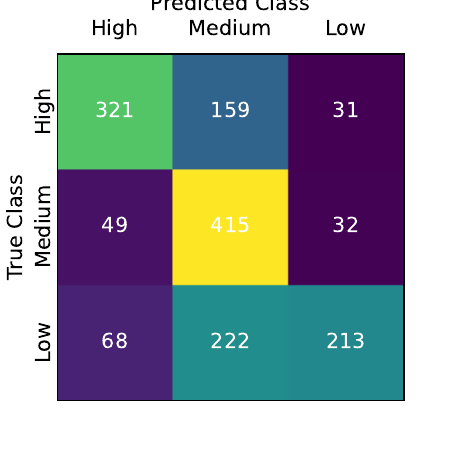}};

        \draw (6.3,-11.1) node[anchor=west]{\rotatebox{90}{\texttt{product}}};

        \draw (1.5,-12.7) node[anchor=north]{\textbf{TF-IDF-LR}};
        \draw (4.7,-12.7) node[anchor=north]{\textbf{GPT-CICLe}};
    \end{tikzpicture}

    \caption{Confusion Matrices for all four prediction tasks. Classes were grouped by class support as shown in Figure~\ref{fig:label_distribution}.}
    \label{fig:confusion_matrices}
\end{figure}
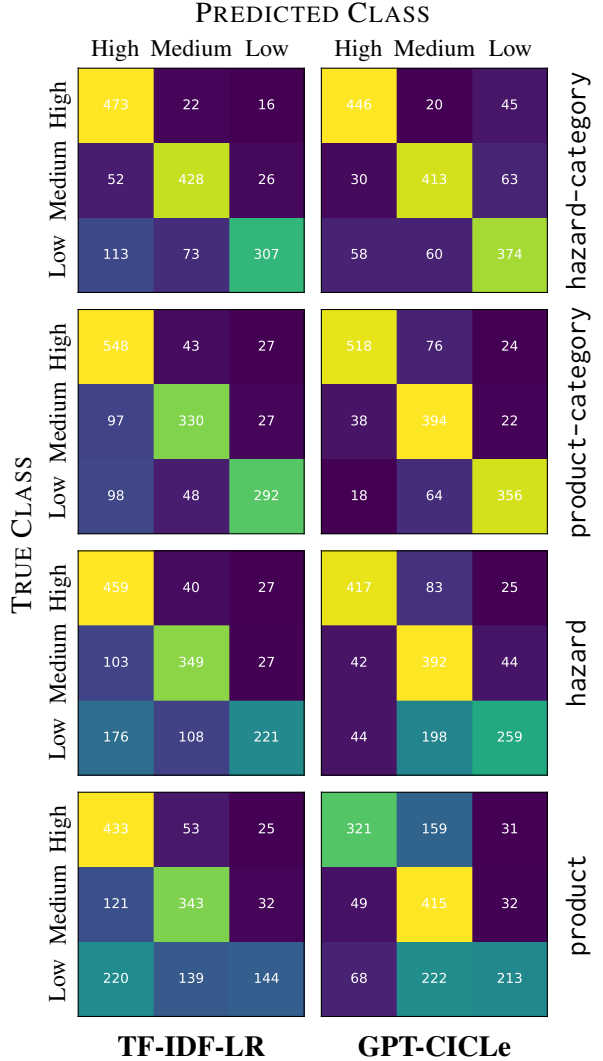

In order to further analyze how prompting performs compared to other classifiers, we plotted confusion matrices of the best non-prompting-classifier (TF-IDF-LR) and GPT-CICLe in Figure~\ref{fig:confusion_matrices}. To achieve a better representation, we grouped classes based on the support zones introduced in the paper.

The figure shows, that while GPT-CICLe is generally worse than TF-IDF-LR in correctly predicting classes with high support, it performs much better on classes with low support. The latter are usually a problem for ML classifiers, as the impact of the few samples that are seen during training is usually overruled by the more supported classes.

TF-IDF-LR is generally more likely to wrongly predict samples to classes with higher support. This is natural, as we do not apply rebalancing techniques during training. For GPT-CICLe, however, this behavior is less evident. We surmise, that the LLM's lower need for training samples is able to balance out this shortcoming of TF-IDF-LR which is used as a basis of GPT-CICLe.

\begin{table}[ht]
    \resizebox*{\linewidth}{!}{
\begin{tabular}{|r|cc|cc|}

            \hline

            \cellcolor{gray!25} &
            \multicolumn{2}{c|}{\cellcolor{gray!25}\textbf{English}} &
            \multicolumn{2}{c|}{\cellcolor{gray!25}\textbf{German}} \\

            \multicolumn{1}{|c|}{\cellcolor{gray!25}\multirow{-2}*{\textbf{Model}}} &
            \cellcolor{gray!25}{\footnotesize $F_1$ (\textit{micro})} &
            \cellcolor{gray!25}{\footnotesize $F_1$ (\textit{micro})} &
            \cellcolor{gray!25}{\footnotesize $F_1$ (\textit{micro})} &
            \cellcolor{gray!25}{\footnotesize $F_1$ (\textit{macro})} \\

            \hline
            \hline

            \multicolumn{2}{|l}{\cellcolor{gray!10}\footnotesize\texttt{hazard-category}} & 
            \multicolumn{3}{r|}{\cellcolor{gray!10}\footnotesize\textit{$10$ classes}} \\

            \hline

            RoBERTa &
            \cellcolor{blue!15}\footnotesize $\bf 0.83$ \tiny $\bf\pm 0.01$ & \cellcolor{blue!15}\footnotesize $\bf 0.57$ \tiny $\bf\pm 0.03$ &
            \cellcolor{blue!15}\footnotesize $\bf 0.39$ \tiny $\bf\pm 0.03$ & \footnotesize $0.13$ \tiny $\pm 0.02$ \\
            
            XLM-R &
            \footnotesize $0.82$ \tiny $\pm 0.01$ & \footnotesize $0.53$ \tiny $\pm 0.02$ &
            \footnotesize $0.37$ \tiny $\pm 0.06$ & \cellcolor{blue!15}\footnotesize $\bf 0.16$ \tiny $\bf\pm 0.04$ \\

            \hline
            \hline

            \multicolumn{2}{|l}{\cellcolor{gray!10}\footnotesize\texttt{product-category}} & 
            \multicolumn{3}{r|}{\cellcolor{gray!10}\footnotesize\textit{$22$ classes}} \\

            \hline

            RoBERTa &
            \footnotesize $0.56$ \tiny $\pm 0.22$ & \footnotesize $0.31$ \tiny $\pm 0.23$ &
            \footnotesize $0.41$ \tiny $\pm 0.10$ & \footnotesize $0.15$ \tiny $\pm 0.10$ \\
\hline
        \end{tabular}
    }
    \caption{Average model $F_1$ and mean deviation over 5 CV-splits on texts in the two most common languages. Blue cells highlight the best score per column and language. For \texttt{hazard} and \texttt{product} we report performance on well-supported classes only.}
    \label{tab:langauge_based_assessment}
\end{table}

\begin{table*}[h]
\begin{center}
\resizebox*{\textwidth}{!}{
\begin{tabular}{|r|cccc|cccc|cccc|}

    \hline

    \cellcolor{gray!25} &
    \multicolumn{4}{c|}{\cellcolor{gray!25}\textbf{Scores} (\textit{\footnotesize all classes})} &
    \multicolumn{4}{c|}{\cellcolor{gray!25}\textbf{Scores} ({\footnotesize $\mathcal{C}_{high}$})} &
    \multicolumn{4}{c|}{\cellcolor{gray!25}\textbf{Scores} ({\footnotesize $\mathcal{C}_{low}$})} \\

    \multicolumn{1}{|c|}{\cellcolor{gray!25}\multirow{-2}*{\textbf{Model}}} &
    \cellcolor{gray!25}{\footnotesize $F_1$ (\textit{micro})} &
    \cellcolor{gray!25}{\footnotesize $F_1$} &
    \cellcolor{gray!25}{\footnotesize Precision} &
    \cellcolor{gray!25}{\footnotesize Recall} &
    \cellcolor{gray!25}{\footnotesize $F_1$ (\textit{micro})} &
    \cellcolor{gray!25}{\footnotesize $F_1$} &
    \cellcolor{gray!25}{\footnotesize Precision} &
    \cellcolor{gray!25}{\footnotesize Recall} &
    \cellcolor{gray!25}{\footnotesize $F_1$ (\textit{micro})} &
    \cellcolor{gray!25}{\footnotesize $F_1$} &
    \cellcolor{gray!25}{\footnotesize Precision} &
    \cellcolor{gray!25}{\footnotesize Recall} \\

    \hline
    \hline

    \multicolumn{7}{|l}{\cellcolor{gray!10}\footnotesize\texttt{hazard-category}} & 
    \multicolumn{6}{r|}{\cellcolor{gray!10}\footnotesize\textit{$10$ classes}} \\

    \hline

    RANDOM &
    \footnotesize $\bf 0.34$ \tiny $\bf\pm 0.01$ & \footnotesize $\bf 0.08$ \tiny $\bf\pm 0.00$ & \footnotesize $\bf 0.07$ \tiny $\bf\pm 0.00$ & \footnotesize $\bf 0.10$ \tiny $\bf\pm 0.00$ &
    \footnotesize $0.50$ \tiny $\pm 0.01$ & \footnotesize $0.33$ \tiny $\pm 0.00$ & \footnotesize $0.50$ \tiny $\pm 0.00$ & \footnotesize $0.25$ \tiny $\pm 0.00$ &
    \footnotesize $\bf 0.00$ \tiny $\bf\pm 0.00$ & \footnotesize $\bf 0.00$ \tiny $\bf\pm 0.00$ & \footnotesize $\bf 0.00$ \tiny $\bf\pm 0.00$ & \footnotesize $\bf 0.00$ \tiny $\bf\pm 0.00$ \\
\hline

\end{tabular}
}
\caption{Average model $F_1$(\textit{micro} \& \textit{macro}), Precision (\textit{macro}), and Recall (\textit{macro}) as well as mean deviation over 5 CV-splits. Blue cells highlight the best score per column and classification task. Bold scores are the best per column, classification task, and type of classifier (\textit{naive, traditional, encoder-transformer, and prompting}).}
\label{tab:results_extended}
\end{center}
\end{table*}

\begin{table*}[h]
\begin{center}
\resizebox*{\textwidth}{!}{
\begin{tabular}{|r|cccc|cccc|cccc|}

    \hline

    \cellcolor{gray!25} &
    \multicolumn{4}{c|}{\cellcolor{gray!25}\textbf{Scores} (\textit{\footnotesize all classes})} &
    \multicolumn{4}{c|}{\cellcolor{gray!25}\textbf{Scores} ({\footnotesize $\mathcal{C}_{high}$})} &
    \multicolumn{4}{c|}{\cellcolor{gray!25}\textbf{Scores} ({\footnotesize $\mathcal{C}_{low}$})} \\

    \multicolumn{1}{|c|}{\cellcolor{gray!25}\multirow{-2}*{\textbf{Model}}} &
    \cellcolor{gray!25}{\footnotesize $F_1$ (\textit{micro})} &
    \cellcolor{gray!25}{\footnotesize $F_1$ (\textit{macro})} &
    \cellcolor{gray!25}{\footnotesize Precision} &
    \cellcolor{gray!25}{\footnotesize Recall} &
    \cellcolor{gray!25}{\footnotesize $F_1$ (\textit{micro})} &
    \cellcolor{gray!25}{\footnotesize $F_1$ (\textit{macro})} &
    \cellcolor{gray!25}{\footnotesize Precision} &
    \cellcolor{gray!25}{\footnotesize Recall} &
    \cellcolor{gray!25}{\footnotesize $F_1$ (\textit{micro})} &
    \cellcolor{gray!25}{\footnotesize $F_1$ (\textit{macro})} &
    \cellcolor{gray!25}{\footnotesize Precision} &
    \cellcolor{gray!25}{\footnotesize Recall} \\

    \hline
    \hline

    \multicolumn{7}{|l}{\cellcolor{gray!10}\footnotesize\texttt{hazard-category}} & 
    \multicolumn{6}{r|}{\cellcolor{gray!10}\footnotesize\textit{$10$ classes; $\alpha=0.05$}} \\

    \hline

    RANDOM &
    \footnotesize $0.33$ & \footnotesize $0.08$ & \footnotesize $0.07$ & \footnotesize $0.10$ &
    \footnotesize $0.51$ & \footnotesize $0.34$ & \footnotesize $0.50$ & \footnotesize $0.25$ &
    \footnotesize $0.00$ & \footnotesize $0.00$ & \footnotesize $0.00$ & \footnotesize $0.00$ \\
    
    MAJORITY &
    \footnotesize $0.34$ & \footnotesize $0.05$ & \footnotesize $0.03$ & \footnotesize $0.10$ &
    \footnotesize $1.00$ & \footnotesize $1.00$ & \footnotesize $1.00$ & \footnotesize $1.00$ &
    \footnotesize $0.00$ & \footnotesize $0.00$ & \footnotesize $0.00$ & \footnotesize $0.00$ \\
\hline

\end{tabular}
}
\caption{Model $F_1$(\textit{micro} \& \textit{macro}), Precision (\textit{macro}), and Recall (\textit{macro}) on 20\% holdout (\textit{i.e. only calculated on the first CV-split to save resources}). Blue cells highlight the best score per column and category.}
\label{tab:results_prompting_extended}
\end{center}
\end{table*}

\begin{table*}[h]
\begin{center}
\resizebox*{\textwidth}{!}{
\begin{tabular}{|r|cccc|cccc|cccc|}

    \hline

    \cellcolor{gray!25} &
    \multicolumn{4}{c|}{\cellcolor{gray!25}\textbf{Scores} (\textit{\footnotesize all classes})} &
    \multicolumn{4}{c|}{\cellcolor{gray!25}\textbf{Scores} ({\footnotesize $\mathcal{C}_{high}$})} &
    \multicolumn{4}{c|}{\cellcolor{gray!25}\textbf{Scores} ({\footnotesize $\mathcal{C}_{low}$})} \\

    \multicolumn{1}{|c|}{\cellcolor{gray!25}\multirow{-2}*{\textbf{Model}}} &
    \cellcolor{gray!25}{\footnotesize $F_1$ (\textit{micro})} &
    \cellcolor{gray!25}{\footnotesize $F_1$ (\textit{macro})} &
    \cellcolor{gray!25}{\footnotesize Precision} &
    \cellcolor{gray!25}{\footnotesize Recall} &
    \cellcolor{gray!25}{\footnotesize $F_1$ (\textit{micro})} &
    \cellcolor{gray!25}{\footnotesize $F_1$ (\textit{macro})} &
    \cellcolor{gray!25}{\footnotesize Precision} &
    \cellcolor{gray!25}{\footnotesize Recall} &
    \cellcolor{gray!25}{\footnotesize $F_1$ (\textit{micro})} &
    \cellcolor{gray!25}{\footnotesize $F_1$ (\textit{macro})} &
    \cellcolor{gray!25}{\footnotesize Precision} &
    \cellcolor{gray!25}{\footnotesize Recall} \\

    \hline
    \hline
    
    \multicolumn{7}{|l}{\cellcolor{gray!10}\footnotesize\texttt{hazard-category}} & 
    \multicolumn{6}{r|}{\cellcolor{gray!10}\footnotesize\textit{$10$ classes; $\alpha=0.05$}} \\

    \hline

    RANDOM &
    \footnotesize $0.33$ & \footnotesize $0.08$ & \footnotesize $0.07$ & \footnotesize $0.10$ &
    \footnotesize $0.51$ & \footnotesize $0.34$ & \footnotesize $0.50$ & \footnotesize $0.25$ &
    \footnotesize $0.00$ & \footnotesize $0.00$ & \footnotesize $0.00$ & \footnotesize $0.00$ \\
    
    MAJORITY &
    \footnotesize $0.34$ & \footnotesize $0.05$ & \footnotesize $0.03$ & \footnotesize $0.10$ &
    \footnotesize $1.00$ & \footnotesize $1.00$ & \footnotesize $1.00$ & \footnotesize $1.00$ &
    \footnotesize $0.00$ & \footnotesize $0.00$ & \footnotesize $0.00$ & \footnotesize $0.00$ \\
\hline

\end{tabular}
}
\caption{Model $F_1$(\textit{micro} \& \textit{macro}), Precision (\textit{macro}), and Recall (\textit{macro}) on 20\% holdout (\textit{i.e. only calculated on the first CV-split to save resources}). Blue cells highlight the best score per column and category. Rows named "MIN-\texttt{[METHOD]}" sigignify a lower bound on performance for prompting with \texttt{[METHOD]}, while rows named "MAX-\texttt{[METHOD]}" signify an upper bound.}
\label{tab:results_prompting_limits}
\end{center}
\end{table*}

\end{document}